\documentclass[10pt,twocolumn,letterpaper]{article}

\usepackage{cvpr}
\usepackage{times}
\usepackage{epsfig}
\usepackage{graphicx}
\usepackage{subfigure}
\usepackage{amsmath}
\usepackage{amssymb}

\usepackage{amsmath,amsfonts,bm}

\def\eqref#1{equation~\ref{#1}}

\def\1{\bm{1}}

\def\vtheta{{\bm{\theta}}}

\def\vv{{\bm{v}}}

\def\vx{{\bm{x}}}

\def\mS{{\bm{S}}}

\DeclareMathAlphabet{\mathsfit}{\encodingdefault}{\sfdefault}{m}{sl}
\SetMathAlphabet{\mathsfit}{bold}{\encodingdefault}{\sfdefault}{bx}{n}

\def\sM{{\mathbb{M}}}

\def\sX{{\mathbb{X}}}

\usepackage{mathtools}
\usepackage{xcolor}
\usepackage{bbding}
\usepackage{pifont}
\usepackage{multirow}
\usepackage{amsthm}
\usepackage{enumitem}

\usepackage{algorithm}
\usepackage{listings}

\pdfoutput=1
\usepackage{pdfpages}
\usepackage{etoolbox}
\makeatletter
\AfterEndEnvironment{algorithm}{\let\@algcomment\relax}
\AtEndEnvironment{algorithm}{\kern2pt\hrule\relax\vskip3pt\@algcomment}
\let\@algcomment\relax
\newcommand\algcomment[1]{\def\@algcomment{\footnotesize#1}}
\renewcommand\fs@ruled{\def\@fs@cfont{\bfseries}\let\@fs@capt\floatc@ruled
  \def\@fs@pre{\hrule height.8pt depth0pt \kern2pt}%
  \def\@fs@post{}%
  \def\@fs@mid{\kern2pt\hrule\kern2pt}%
  \let\@fs@iftopcapt\iftrue}
\makeatother

\newtheorem{lemma}{Lemma}

\newcommand*{\defeq}{\mathrel{\vcenter{\baselineskip0.5ex \lineskiplimit0pt
			\hbox{\scriptsize.}\hbox{\scriptsize.}}}%
	=}

\definecolor{citecolor}{rgb}{0.133, 0.752, 0.133}
\usepackage[pagebackref=false,breaklinks=true,letterpaper=true,colorlinks,citecolor=citecolor,bookmarks=false]{hyperref}

\definecolor{Highlight}{HTML}{39b54a}  

\newlength\savewidth\newcommand\shline{\noalign{\global\savewidth\arrayrulewidth
  \global\arrayrulewidth 1pt}\hline\noalign{\global\arrayrulewidth\savewidth}}
\newcommand{\tablestyle}[2]{\setlength{\tabcolsep}{#1}\renewcommand{\arraystretch}{#2}\centering\footnotesize}

\newcommand\blfootnote[1]{%
  \begingroup
  \renewcommand\thefootnote{}\footnote{#1}%
  \addtocounter{footnote}{-1}%
  \endgroup
}

\cvprfinalcopy 

\ifcvprfinal\fi
\begin{document}



\title{Cross-Batch Memory for Embedding Learning}

	\author{Xun Wang$^*$, Haozhi Zhang$^*$ , Weilin Huang$^\dagger$, Matthew R. Scott\\
		Malong Technologies\\
		{\tt\small \{xunwang, haozhang, whuang, mscott\}@malong.com}
	}
\maketitle

\thispagestyle{empty}

\begin{abstract}
\begin{figure*}[t]
\vspace{0.em}
\centering
\centering
\includegraphics[width=0.3\textwidth]{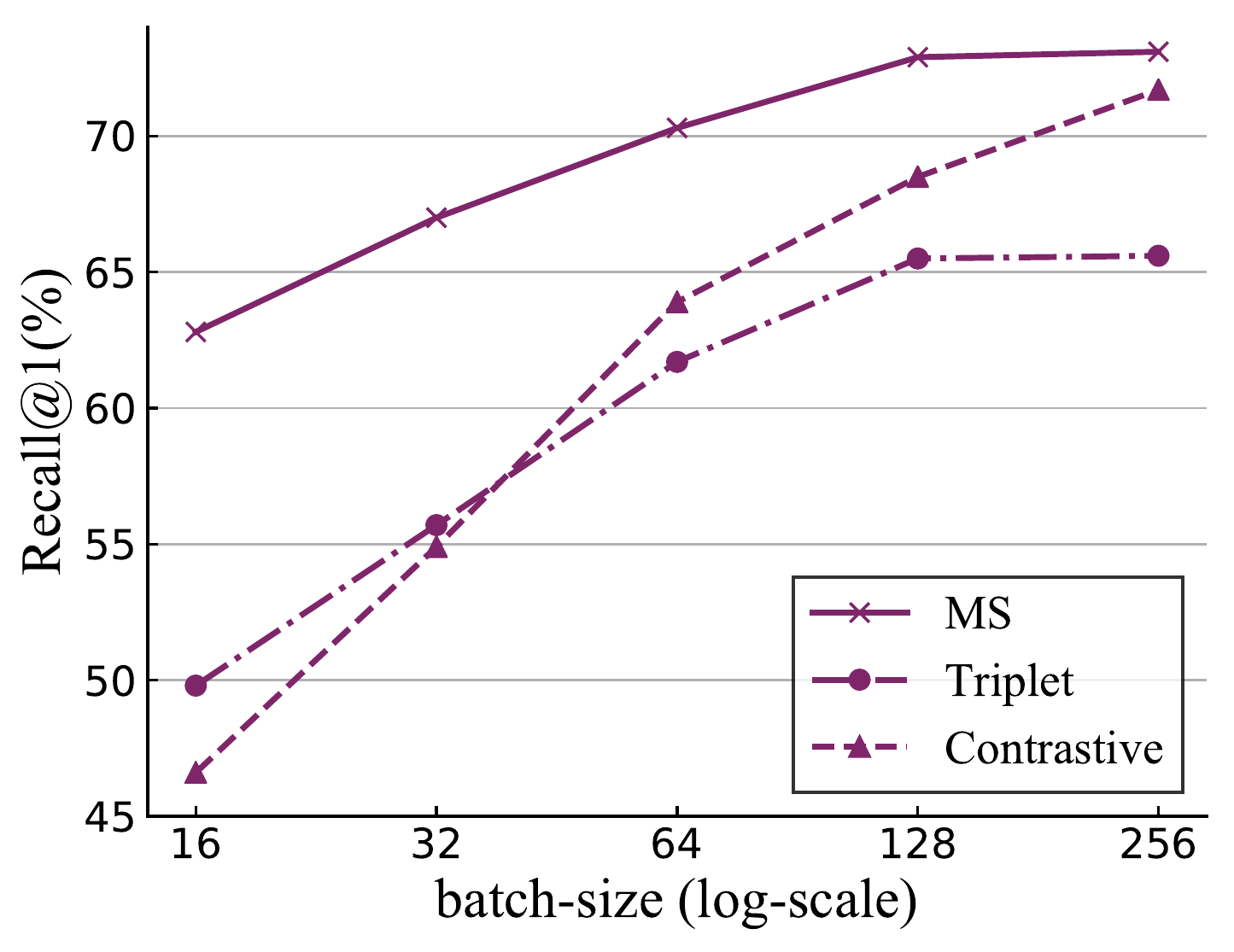}
\hfill
\centering
\includegraphics[width=0.3\textwidth]{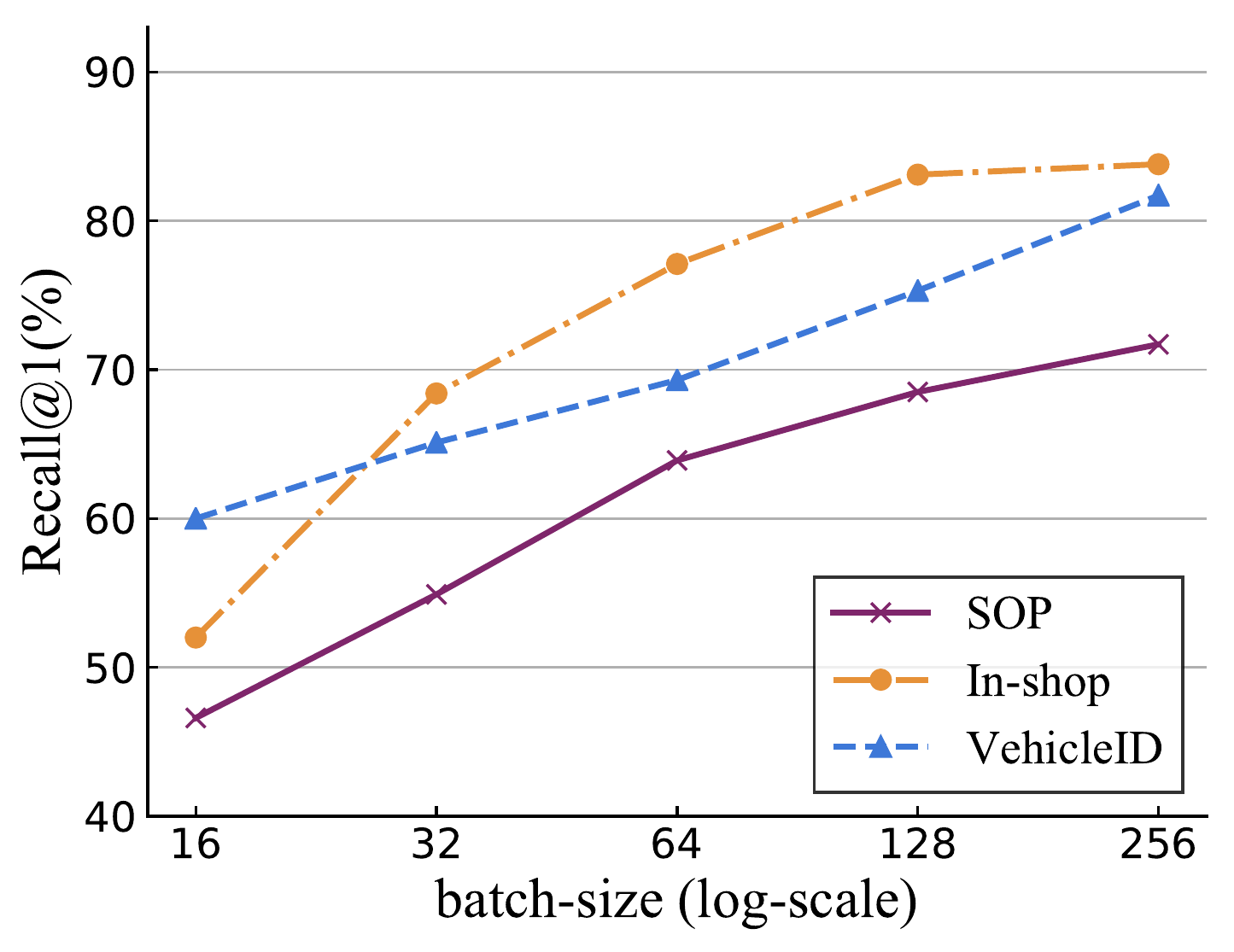}
\hfill
\centering
\includegraphics[width=0.3\textwidth]{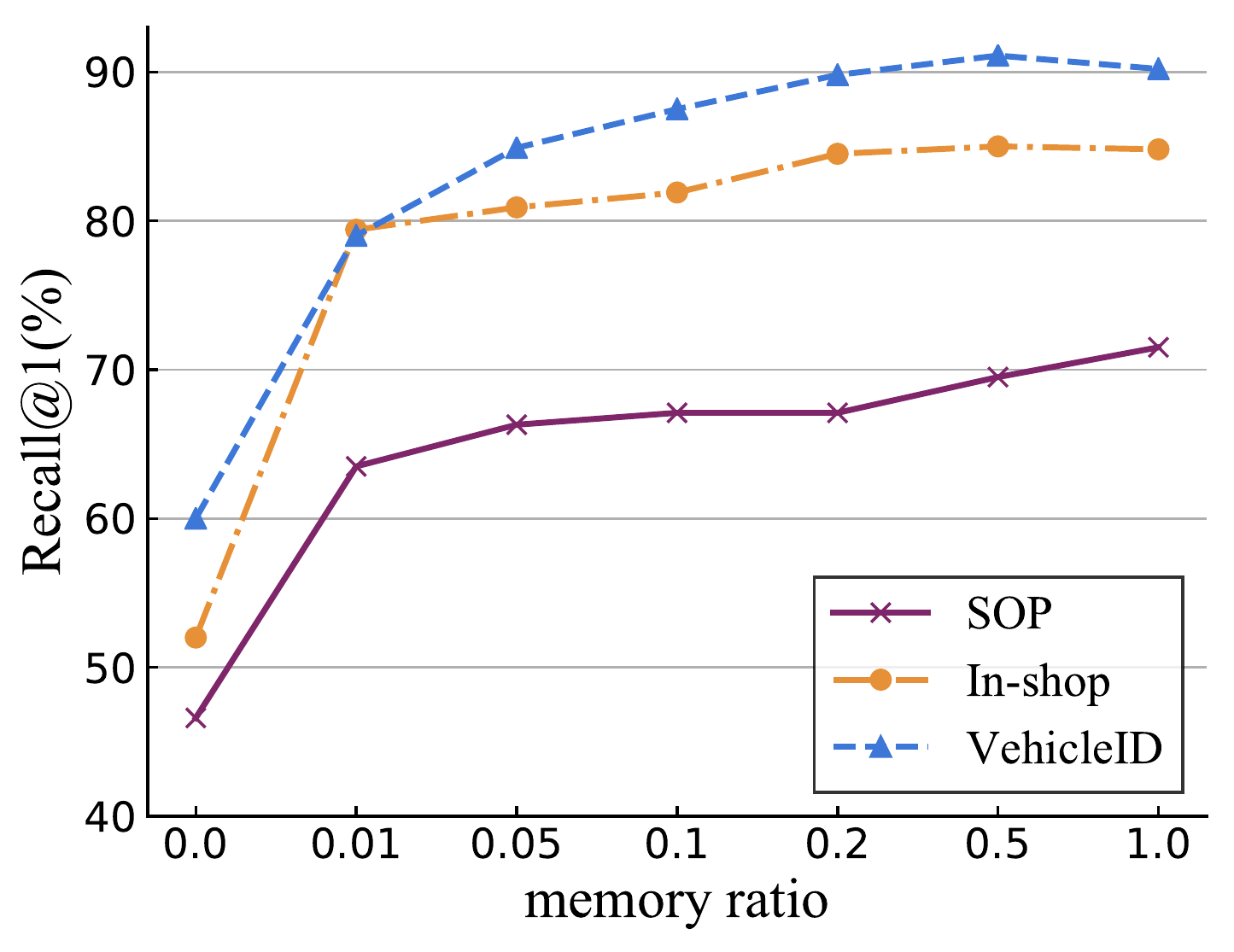}

\caption{\small R@1 results with \textbf{ GoogleNet}. \textbf{Left:} R@1 on \textbf{SOP} {$vs.$} mini-batch size with contrastive, triplet and MS approaches. \textbf{Middle:}  R@1 {$vs.$} mini-batch size by varying datasets. \textbf{Right} R@1 {$vs.$} memory ratio at mini-batch size 16 with contrastive loss.} 
\label{fig:bs-mem}
\vspace{-1em}
\end{figure*}
Mining informative negative instances are of central importance to deep metric learning (DML), however this task is intrinsically limited by mini-batch training, where only a mini-batch of instances is accessible at each iteration.
In this paper, we identify a {``slow drift"} phenomena by observing that the embedding features drift exceptionally slow even as the model parameters are updating throughout the training process. 
This suggests that the features of instances computed at preceding iterations can be used to considerably approximate their features extracted by the current model.
We propose a cross-batch memory (XBM) mechanism that memorizes the embeddings of past iterations, allowing the model to collect sufficient hard negative pairs across multiple mini-batches - even over the whole dataset. Our XBM can be directly integrated into a general pair-based DML framework, where the XBM augmented DML can boost performance considerably. In particular, without bells and whistles, a simple contrastive loss with our XBM can have large R@1 improvements of 12\%-22.5\% on three large-scale image retrieval datasets, surpassing the most sophisticated state-of-the-art methods \cite{wang2019multi, roth2019mic, cakir2019deep}, by a large margin. Our XBM is conceptually simple, easy to implement - using several lines of codes, and is memory efficient - with a negligible 0.2 GB extra GPU memory. Code is available at: \url{https://github.com/MalongTech/research-xbm}.

\blfootnote{$^*$Equal contribution $^\dagger$Corresponding author}

\end{abstract}

\section{Introduction}
Deep metric learning (DML) aims to learn an embedding space where instances from the same class are encouraged to be closer than those from different classes. 
As a fundamental problem in computer vision, DML has been applied to various tasks, including image retrieval \cite{wohlhart2015learning,He_2018_CVPR,Grabner_2018_CVPR}, face recognition \cite{Wen2016},
zero-shot learning \cite{zhang2016zero,bucher2016improving,Yelamarthi_2018_ECCV}, visual tracking \cite{leal2016learning,tao2016siamese} and person re-identification \cite{Yu_2018_ECCV,in-defense}.

A family of DML approaches are known as pair-based, whose objectives can be defined in terms of pair-wise similarities within a mini-batch, such as contrastive loss \cite{contrass1}, triplet loss \cite{facenet}, lifted-structure loss\cite{lifted-structured-loss}, n-pairs loss \cite{n-pairs}, multi-similarity (MS) loss \cite{wang2019multi} and \etc. Moreover, most existing pair-based DML methods can be unified as weighting schemes under a General Pair Weighting (GPW) framework \cite{wang2019multi}.
The performance of pair-based methods heavily rely on their capability of mining informative negative pairs.  To collect sufficient informative negative pairs from each mini-batch, many efforts have been devoted to improving the sampling schemes, which can be categorized into two main directions: (1) sampling informative mini-batches based on global data distribution  \cite{suh2019stochastic,HTL,sanakoyeu2019divide,suh2019stochastic,smart-mining}; (2) weighting informative pairs within each individual mini-batch
\cite{lifted-structured-loss,n-pairs,wang2019multi,histogram,sampling}.

Various sophisticated sampling schemes have been developed, but the hard mining ability is inherently limited by the size of a mini-batch, which the number of possible training pairs depends on. Therefore, to improve the sampling scheme, it is straightforward to enlarge the mini-batch, which can boost the performance of pair-based DML methods immediately. 
We demonstrate by experiments that the performance of pair-based approaches, such as contrastive loss \cite{contrass1} and recent MS loss \cite{wang2019multi}, can be improved strikingly when the mini-batch grows larger on large-scale datasets (Figure~\ref{fig:bs-mem}, left and middle).  It is not surprising because the number of negative pairs grows \emph{quadratically} \wrt the mini-batch size. 
However, simply enlarging a mini-batch is not an ideal solution to solve the hard mining problem due to two limitations: (1) the mini-batch size is limited by the GPU memory and computational cost; (2) a large mini-batch (\eg 1800  used in \cite{facenet}) often requires cross-device synchronization, which is a challenging engineering task. 
A naive solution to collect abundant informative pairs is to compute the features of instances in the whole training set at each training iteration, and then search for hard negative pairs from the whole dataset.
Obviously, this solution is extremely time consuming, especially for a large-scale dataset, but it inspired us to \textit{break the limit of mining hard negatives within a single mini-batch}.

In this paper, we identify an interesting {\it ``slow drift''} phenomena that the embedding of an instance actually drifts at a relatively slow rate throughout the training process. It suggests that the deep features of a mini-batch computed at past iterations can considerably approximate to those extracted by current model. Based on the {\it ``slow drift''} phenomena, we propose a cross-batch memory (\textbf{XBM}) module to record and update the deep features of recent mini-batches, allowing for mining informative examples \textit{across multiple  mini-batches}.  Our XBM can provide plentiful hard negative pairs by directly connecting each anchor in the current mini-batch with the embeddings from recent mini-batches.

Our XBM is conceptually simple, easy to implement and memory efficient. The memory module can be updated using a simple enqueue-dequeue mechanism by leveraging the computation-free features computed at the past iterations, with only about a negligible 0.2 GB of extra GPU memory utilized. More importantly, our XBM can be directly integrated into most existing pair-based methods with just several lines of codes, and can boost performance considerably.
We evaluate our XBM with various conventional pair-based DML techniques on three widely used large-scale image retrieval datasets: Stanford Online Products (SOP) \cite{lifted-structured-loss}, In-shop Clothes Retrieval (In-shop) \cite{DeepFashion}, and PKU VehicleID (VehicleID) \cite{liu2016deep}. 
In Figure~\ref{fig:bs-mem} (middle and right), our approach demonstrates excellent robustness and brings consistent performance improvements across all settings; under the same configurations, our XBM obtains extraordinary R@1 improvements on all three datasets compared with the corresponding pair-based methods (\eg over 20\% for contrastive loss). 
Furthermore, with our XBM, a simple contrastive loss can easily outperform the most state-of-the-art sophisticated methods, such as \cite{wang2019multi, roth2019mic, cakir2019deep}, by a large margin. 

In parallel to our work, He \etal \cite{he_2019_moco} built a dynamic dictionary as a queue of preceding mini-batches to provide a rich set of negative samples for unsupervised learning (with a contrastive loss). However, unlike \cite{he_2019_moco} which uses a specific encoding network to compute the features of current mini-batch, our features are computed more efficiently by taking them directly from the forward of the current model with no additional computational cost. More importantly, to solve the problem of feature drift, He \etal designed a momentum update that slowly progresses the key encoder to ensure the consistency between different iterations, while we identify the \emph{``slow drift''} phenomena which suggests that the features can become stable by themselves when the early phase of training finishes.
\section{Related Work}
\noindent \textbf{Pair-based DML.} 
 Pair-based DML methods can be optimized by computing the pair-wise similarities between instances in the embedding space \cite{contrastive, lifted-structured-loss, facenet, histogram, n-pairs, wang2019multi}. 
Contrastive loss \cite{contrastive} is one of the classic pair-based DML methods, which learns a discriminative metric via Siamese networks.
It encourages the deep features of positive pairs to be closer to each other, and those of negative pairs to be farther than a fixed threshold. Triplet loss \cite{facenet} requires the similarity of a positive pair to be higher than that of a negative pair (with the same anchor) by a given margin. 

Inspired by contrastive loss and triplet loss, a number of pair-based DML algorithms have been developed, which attempted to weight \emph{all pairs in a mini-batch}, such as up-weighting informative pairs  (\eg N-pair loss \cite{n-pairs}, Multi-Similarity (MS) loss \cite{wang2019multi}) through a log-exp formulation, or sampling negative pairs uniformly \wrt pair-wise distance \cite{sampling}. Generally, pair-based methods can be cast into a unified weighting formulation through General Pair Weighting (GPW) framework \cite{wang2019multi}. 

However, most deep models are trained with SGD where only a mini-batch of samples are accessible at each iteration, and the size of a mini-batch can be relatively small compared to the whole dataset, especially for a large-scale dataset. Moreover, a large fraction of the pairs is less informative as the model learns to embed most trivial pairs correctly. Thus the conventional pair-based DML techniques suffer from lacks of hard negative pairs, which are critical to promote model training. 

To alleviate the aforementioned problems, a number of approaches have been developed to collect more potential information contained in a mini-batch, such as building a class-level hierarchical tree \cite{HTL}, updating class-level signatures to select hard negative instances \cite{suh2019stochastic}, or obtaining samples from an individual cluster \cite{sanakoyeu2019divide}. Unlike these approaches which aim to enrich the information in a mini-batch, our XBM are designed to directly mine hard negative examples across multiple mini-batches.\\

\noindent \textbf{Proxy-based DML.} 
There is another branch of DML methods aiming to learn the embeddings by comparing each sample with proxies, including proxy NCA \cite{proxyloss}, NormSoftmax \cite{zhaiclassification} and SoftTriple \cite{qian2019softtriple}. In fact, our XBM module can be regarded as the proxies to some extent. However, there are two main differences between the proxy-based methods and our XBM module: (1) proxies are often optimized along with the model weights, while the embeddings of our memory are directly taken from the past mini-batches; (2) proxies are used to represent the class-level information, whereas the embedding of our memory computes the information for each individual instance. 
Both proxy-based methods and our XBM augmented pair-based methods are able to capture the global distribution of data over the whole dataset during training. \\

\noindent \textbf{Feature Memory Module.}
Non-parametric memory module for embedding learning has shown power in various computer visual tasks \cite{vinyals2016matching,xiao2017joint,wu2018improving,wu2018unsupervised,zhong2019invariance,Li_Memory_2019_ICCV}.
 For examples, the external memory can be used to address the unaffordable computational demand of conventional NCA \cite{wu2018improving} in large-scale recognition, and encourage instance-invariance in domain adaptation \cite{zhong2019invariance,wu2018unsupervised}. But only the positive pairs are optimized, while the negatives are ignored in \cite{wu2018improving}. Our XBM is able to to provide a rich set of negative examples for the pair-based DML methods, which is more generalized and can make full use of the past embeddings.
 The key distinction is that existing memory modules only store the embeddings of current mini-batch \cite{vinyals2016matching}, or maintain the whole dataset \cite{wu2018improving,zhong2019invariance} with a moving average update, while our XBM is maintained as a dynamic queue of mini-batches, which is more flexible and applicable in extremely large-scale datasets. 

\section{Cross-Batch Memory Embedding Networks}

\begin{figure*}[t]
\vspace{0em}
\centering
\includegraphics[width=0.97\textwidth, trim=30 250 50 150, clip]{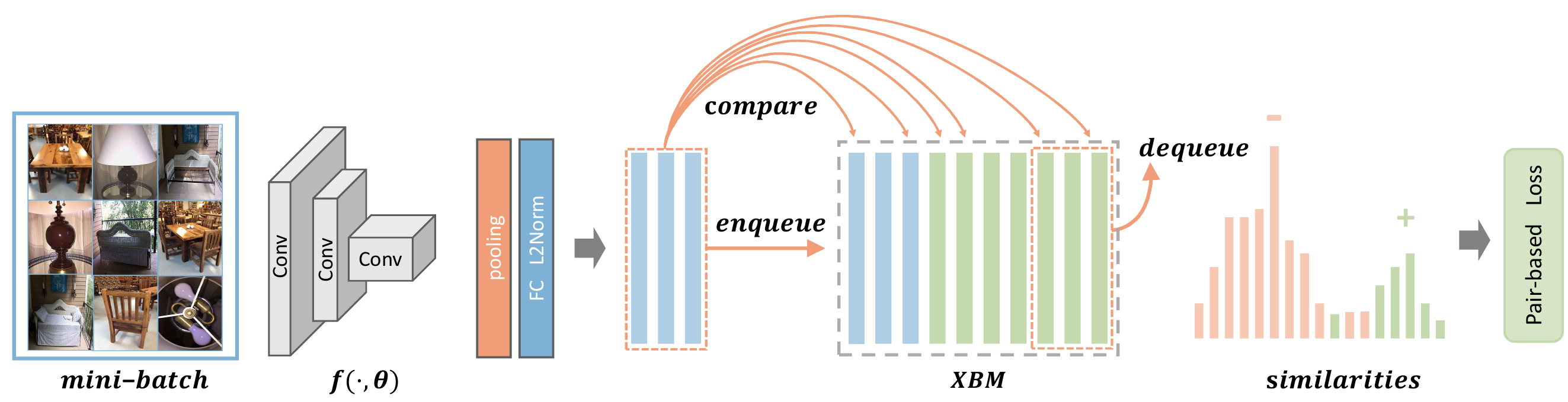}

\caption{\small \textbf{\small Cross-Batch Memory (XBM)} trains an embedding network by comparing each anchor with a memory bank using a pair-based loss. The memory bank is maintained as a queue with the current mini-batch enqueued and the oldest mini-batch dequeued. Our XBM enables a large amount of valid negatives for each anchor to benefit the model training with many pair-based methods.}
\label{fig:main}
\vspace{-1em}
\end{figure*}

In this section, we first analyze the limitation of existing pair-based DML methods. Then we introduce the {\it``slow drift"} phenomena, which provides the underlying evidence that supports our cross-batch mining approach. Finally, we describe our XBM module and integrate it into existing pair-based DML methods.

\subsection{Delving into Pair-based DML}
\label{pair-based}

Let $\sX=\{\vx_1, \vx_2, \dots, \vx_N\}$ denotes a set of training instances, and $y_i$ is the corresponding label of $\vx_i$. An embedding function, $f(\cdot; \vtheta)$, projects a data point $\vx_i$ onto a $D$-dimensional unit hyper-sphere, $\vv_i = f(\vx_i; \vtheta)$. We measure the similarity between two instances of a pair in the embedding space. During training, we denote an affinity matrix of all pairs within the current mini-batch as $\mS$, whose $(i,j)$ element is the cosine similarity between the embeddings of the $i$-th sample and the $j$-th sample: $\vv_i^T\vv_j$.

To facilitate further analysis, we delve into the pair-based DML methods by using the GPW framework described in \cite{wang2019multi}. With GPW, a pair-based function can be formulated in a unified pair-weighting form:
\begin{equation}
\small
\label{weight}
 \mathcal{L}  =  \frac{1}{m} \sum_{i=1}^{m}
\left[\sum_{y_j \neq y_i}^{m} w_{ij}  \mS_{ij} - \sum_{y_j = y_i}^{m} w_{ij} \mS_{ij} \right],
\end{equation}
where $m$ is the mini-batch size, and $w_{ij}$ is the weight assigned to $\mS_{ij}$.
Eq. \ref{weight} shows that any pair-based method can be considered as a weighting scheme focusing on informative pairs. Here we list the weighting schemes of contrastive loss, triplet loss and MS loss.
\begin{itemize}	

\item[--] \textbf{Contrastive loss.}
For each negative pair, $w_{ij}=1$ if $\mS_{ij} > \lambda$, otherwise $w_{ij}=0$. The weights of all positive pairs are set to 1.

\item[--] \textbf{Triplet loss.}
For each negative pair, $w_{ij}= |\mathcal{P}_{ij}|$, where $\mathcal{P}_{ij}$ is the valid positive set sharing the anchor. Formally, $\mathcal{P}_{ij} = \{\vx_{i,k}| y_{k} = y_{i}, \  \text{and} \ \mS_{ik} < \mS_{ij} + \eta\}$ and $\eta$ is a predefined margin in triplet loss. Similarly, we can obtain the triplet weight for a positive pair.

\item[--] \textbf{MS loss.} Unlike contrastive loss and triplet loss which only assigns a weight with integer value, MS loss \cite{wang2019multi} is able to weight the pairs more properly by jointly considering multiple similarities. The MS weight for a negative pair is computed as:
\begin{equation*}
\small
 w_{ij} = \frac{ e^{\beta \left(\mS_{ij} - \lambda \right)} }{1 + \sum\limits_{k \in \mathcal{N}_i} e^{\beta \left(\mS_{ik} - \lambda \right)}},
\end{equation*}
where $\beta$ and $\lambda$ are hyper-parameters, and $\mathcal{N}_i$ is the valid negative set for the anchor $\vx_{i}$. The MS weight for a positive pair can be computed similarly.
\end{itemize}	

In fact, the main path of developing pair-based DML is to design a better weighting mechanism for pairs within a mini-batch. Generally, with a small mini-batch ({\eg 16 or 32}), the sophisticated weighting schemes can perform much better (Figure~\ref{fig:bs-mem}, left). 
However, beyond the weighting scheme, the mini-batch size is also of great importance to DML. 
Figure~\ref{fig:bs-mem} (left and middle) shows the R@1s of various pair-based methods are increased considerably by using a larger mini-batch  on large-scale datasets.
%
Intuitively, the number of negative pairs increase quadratically when the mini-batch size grows, which naturally provides more informative pairs. Instead of developing another sophisticated but highly complicated algorithm to weight the informative pairs, our intuition is to simply collect sufficient informative negative pairs, where a simple weighting scheme, such as contrastive loss, can easily outperform the stage-of-the-art weighting approaches. This provides a new path that is straightforward yet more efficient to solve the hard mining problem in DML.

Naively, a straightforward solution to collect more informative negative pairs is to increase the mini-batch size. However, training deep networks with a large mini-batch is limited by GPU memory, and often requires massive data flow communication between multiple GPUs. To this end, we attempt to achieve the same goal by introducing an alternative approach using very low GPU memory and minimum computation burden.
We propose a XBM module that allows the model to collect informative pairs over multiple past mini-batches, based on the \emph{``slow drift"} phenomena as described below.

\subsection{Slow Drift Phenomena}

The embeddings of past mini-batches are usually considered out-of-date because the model parameters are changing throughout the training process \cite{he_2019_moco,suh2019stochastic,qian2019softtriple}. Such out-of-date features are always discarded, but we learn that they can be an important resource, while being computation-free, by identifying the \emph{``slow drift"} phenomena. We study the drifting speed of embeddings by measuring the difference of features for the same instance computed at different training iterations.
Formally, the feature drift of an input $\vx$ at $t$-th iteration with step $\Delta t$ is defined as:
\begin{equation}
\begin{split}
 D(x, t; \Delta t)& \defeq||f(\vx; \vtheta^t) - f(\vx; \vtheta^{t-\Delta t})||_2^2 \\
\end{split}
\end{equation}

\begin{figure}[t]
\vspace{1em}
\centering
\includegraphics[width=0.47\textwidth, trim=10 10 10 10, clip]{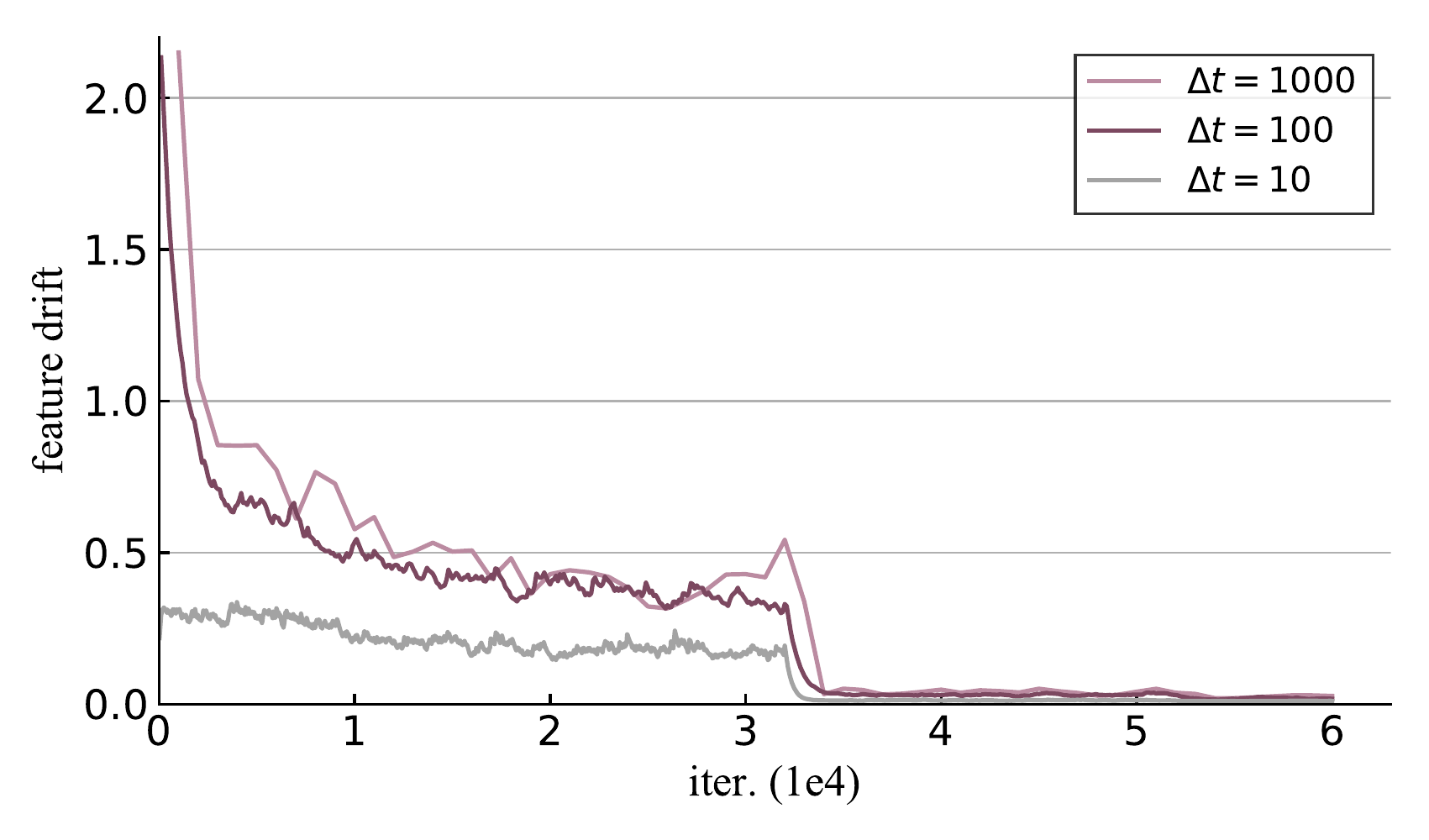}

\caption{\small \textbf{Feature drift} with different steps on SOP. The embeddings of training instances drift within a relatively small distance even under a large interval, \eg $\Delta t=1000$.}
\label{fig:drift}
\vspace{-1em}
\end{figure}

We train GoogleNet \cite{inception} from \emph{scratch} with a contrastive loss, and compute the average feature drift for a set of randomly sampled instances at different steps: $\{10, 100, 1000\}$ (in Figure~\ref{fig:drift}). The feature drift is consistently small, within only \eg 10 iterations.
For the large steps, \eg 100 and 1000, the features change drastically at the early phase, but become relatively stable within about 3K iterations. Furthermore, when the learning rate decreases, the drift gets extremely slow. We define such phenomena as {\it ``slow drift"}, which suggests that with a certain number of training iterations, the embeddings of instances can drift very slowly, resulting in marginal differences between the features computed at different training iterations.

Furthermore, we demonstrate that such {\it ``slow drift"} phenomena can provide a strict upper bound for the error of gradients of a pair-based loss. For simplicity, we consider the contrastive loss of one single negative pair $\mathcal{L} = \vv_i^T \vv_j$, where $\vv_i$, $\vv_j$ are the embeddings of current model and $\widetilde{\vv}_j$ is an approximation of $\vv_j$.

\begin{lemma}
 Assume $|| \vv_j - \widetilde{\vv}_j ||^2_2 <  \epsilon$ \text{,} $\widetilde{\mathcal{L}} = \vv_i^T\widetilde{\vv}_j$ \text{and } $f$ satisfies Lipschitz continuous condition, then the error of gradients related to $\vv_i$ is,
\begin{equation}
\Big|\Big|\frac{\partial\mathcal{L}}{\partial \vtheta} - \frac{\partial \widetilde{\mathcal{L}}}{\partial \vtheta}\Big|\Big|^2_2 < C \epsilon,
\end{equation}
where $C$ is the Lipschitz constant.
\label{lemma}
\end{lemma}

 Proof and discussion of Lemma~\ref{lemma} are provided in Supplementary Materials. Empirically, $C$ is often less than 1 with the backbones used in our experiments. Lemma~\ref{lemma} suggests that the error of gradients is controlled by the error of embeddings under Lipschitz assumption. Thus, the \emph{``slow drift"} phenomenon ensures that mining across mini-batches can provide negative pairs \textbf{with valid information} for pair-based methods.
 
 In addition, we discover that the \emph{``slow drift"} of embeddings is not a special phenomena in DML, and also exists in other conventional tasks, as shown in Supplementary Materials.
 
\subsection{Cross-Batch Memory Module}
\begin{algorithm}[t]
    \caption{Pseudocode of XBM.}
    \label{alg:code}
    \algcomment{\fontsize{7.2pt}{0em}\selectfont
    \vspace{-2.3em}
    }
    \definecolor{codeblue}{rgb}{0.580,0.337,0.447}
    \lstset{
      backgroundcolor=\color{white},
      basicstyle=\fontsize{7.2pt}{7.2pt}\ttfamily\selectfont,
      columns=fullflexible,
      breaklines=true,
      captionpos=b,
      commentstyle=\fontsize{7.2pt}{7.2pt}\color{codeblue},
      keywordstyle=\fontsize{7.2pt}{7.2pt},
    }

\begin{lstlisting}[language=python]
train network f conventionally with K epochs
initialize XBM as queue M 

for x, y in loader:  #  x: data, y: labels
    anchors = f.forward(x) 

    # memory update
    enqueue(M, (anchors.detach(), y))
    dequeue(M)
    
    # compare anchors with M
    sim = torch.matmul(anchors.transpose(), M.feats)
    loss = pair_based_loss(sim, y, M.labels)
    
    loss.backward()
    optimizer.step()
\end{lstlisting}
\end{algorithm}

We first describe our cross-batch memory (XBM) module, with model initialization and updating mechanism. Then we show that our memory module is easy to implement, can be directly integrated into existing pair-based DML framework as a plug-and-play module, by simply using several lines of codes (in Algorithm~\ref{alg:code}). \\

\noindent\textbf{XBM.}
As the feature drift is relatively large at the early epochs, we warm up the neural networks with 1k iterations, allowing the model to reach a certain local optimal field where the embeddings become more stable.
Then we initialize the memory module $\sM$ by computing the features of a set of randomly sampled training images with the warm-up model. Formally,  ${\sM}=\{(\widetilde{\vv}_1, \widetilde{y}_1), (\widetilde{\vv}_2, \widetilde{y}_2), \dots, (\widetilde{\vv}_m, \widetilde{y}_M)\}$, where $\widetilde{\vv}_i$ is initialized as the embedding of the \emph{i}-th sample $\vx_i$, and $M$ is the memory size. We define a \emph{memory ratio} as $R_{\sM} \defeq M/N$, the ratio of memory size to the training size.

We maintain and update our XBM module as a \textit{queue}: at each iteration, we enqueue the embeddings and labels of the current mini-batch, and dequeue the entities of the earliest mini-batch. Thus our memory module is updated with embeddings of the current mini-batch directly, without any additional computation. Furthermore, the whole training set can be cached in the memory module, because very limited memory is required for storing the embedding features, \eg $512$-$d$ float vectors. See the other update strategy in Supplementary Materials. \\

\noindent\textbf{XBM augmented Pair-based DML.} We perform hard negative mining with our XBM on the pair-based DML. For a pair-based loss, based on GPW in \cite{wang2019multi}, it can be cast into a unified weighting formulation of pair-wise similarities within a mini-batch in Eqn.(\ref{weight}), where a similarity matrix is computed within a mini-batch, $\mS$. To perform our XBM mechanism, we simply compute a cross-batch similarity matrix $\widetilde{\mS}$ between the instances of current mini-batch and the memory bank.

Formally, the memory augmented pair-based DML can be formulated as below:
\begin{equation}
\label{memory_pair}
\small
 \mathcal{L}  =  \frac{1}{m} \sum_{i=1}^{m} \mathcal{L}_i = \sum_{i=1}^{m} \left[\sum_{\widetilde{y}_j \neq y_i}^{M} w_{ij}  \widetilde{\mS}_{ij} - \sum_{\widetilde{y}_j= y_i}^{M} w_{ij} \widetilde{\mS}_{ij} \right],
\end{equation}
where $\widetilde{\mS}_{ij} = \vv_i^T \widetilde{\vv}_j$.
The memory augmented pair-based loss in Eqn.(\ref{memory_pair}) is in the same form as the normal pair-based loss in Eqn.(\ref{weight}), by computing a new similarity matrix $\widetilde{\mS}$. Each instance in current mini-batch is compared with all the instances stored in the memory, enabling us to collect sufficient informative pairs for training. The gradient of the loss $\mathcal{L}_i$ \wrt $\vv_i$ is,
\begin{equation}
    \frac{\partial \mathcal{L}_i}{ \partial \vv_i} = \sum_{\widetilde{y}_j \neq y_i}^{M} w_{ij}  \widetilde{\vv}_{j} - \sum_{\widetilde{y}_j= y_i}^{M} w_{ij} \widetilde{\vv}_{j}
\end{equation}
and the gradients \wrt $\vv_i$ model parameters ($\vtheta$) can be computed through a chain rule:
\begin{equation}
    \frac{\partial \mathcal{L}_i}{ \partial \vtheta} = \frac{\partial \mathcal{L}_i}{ \partial \vv_i}
    \frac{\partial \vv_i}{\partial \vtheta}
\end{equation}
Finally, the model parameters $\vtheta$ are optimized through stochastic gradient descent. Lemma~\ref{lemma} ensures that the gradient error raised by embedding drift can be strictly constrained with a bound, which minimizes the side effect to the model training. \\

\begin{figure}[t]
\vspace{-0.6em}
\centering
\includegraphics[width=0.47\textwidth, trim=12 15 0 0, clip]{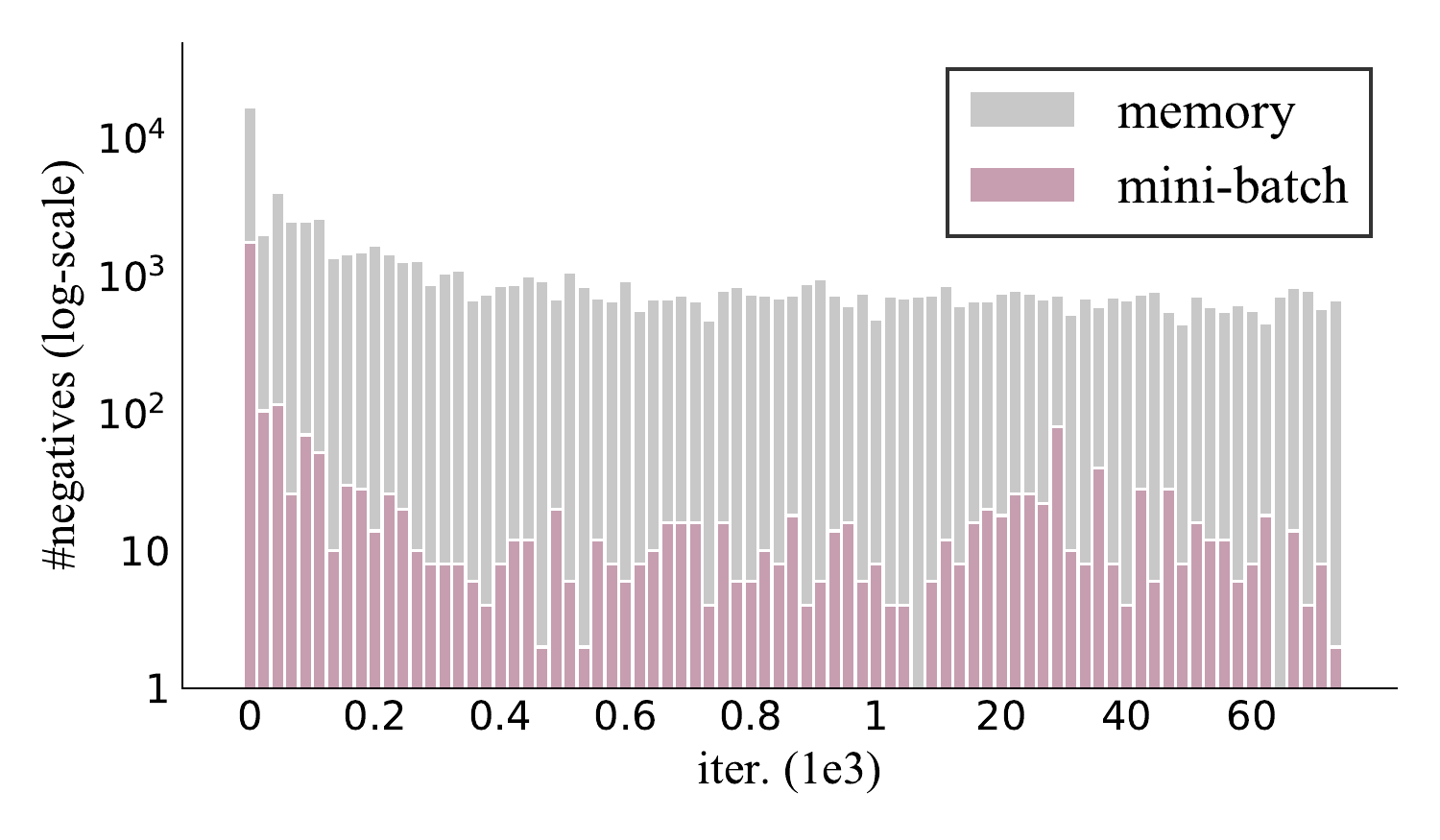}
\caption{\small The number of valid negative examples from mini-batch and that from memory per iteration. Model is trained on SOP with $R_M=1$, mini-batch size 64 and GoogleNet as the backbone.}
\label{fig:negative_count}
\vspace{0em}
\end{figure}

\noindent\textbf{Hard Mining Ability.} 
We investigate the hard mining ability of our XBM mechanism. We study the amount of valid negative pairs produced by our memory module at each iteration. A negative pair with non-zero gradient is considered as valid. The statistical result is illustrated in Figure~\ref{fig:negative_count}.
Throughout the training procedure, our memory module steadily contributes about 1,000 hard negative pairs per iteration, whereas less than 10 valid pairs are generated by the original mini-batch mechanism.

\begin{figure}[t]
\vspace{0em}
\centering
\includegraphics[width=0.47\textwidth, trim=120 100 170 160, clip]{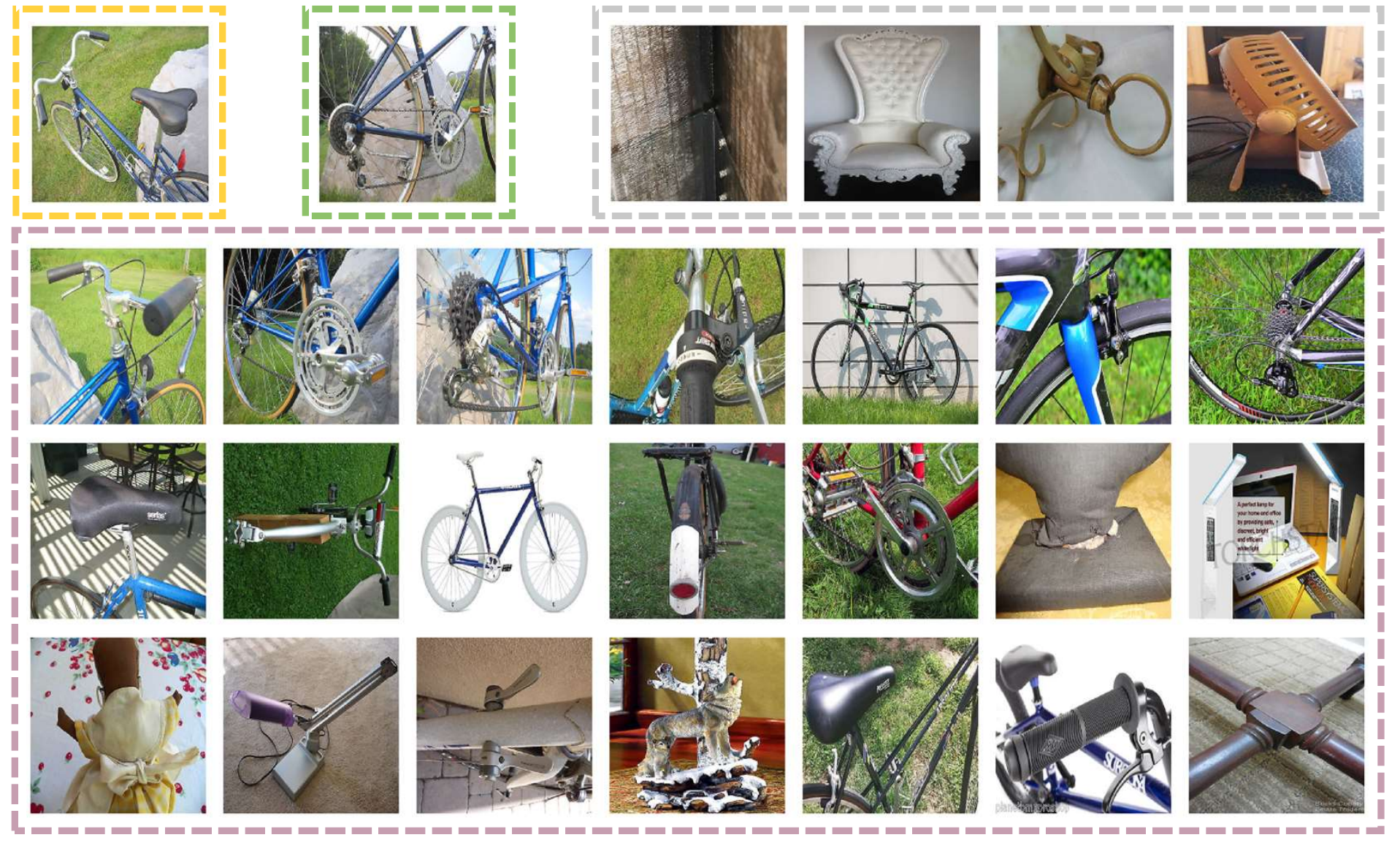}
\caption{\small Given an anchor image ({\color[rgb]{1, 0.819, 0.247} yellow}), examples of positive ({\color[rgb]{0.556, 0.760, 0.415} green}) and negative from mini-batch ({\color[rgb]{0.6, 0.6, 0.6} gray}) and that from memory ({\color[rgb]{0.729, 0.533, 0.623} purple}). Current mini-batch can only bring few valid negatives with less information, while our XBM module can provide a wide variety of informative negative examples.}
\label{fig:negative_show}
\vspace{-1.5em}
\end{figure}

Qualitative hard mining results are shown in Figure \ref{fig:negative_show}. Given a \emph{bicycle} image as an anchor, the mini-batch provides limited and different images, \eg \emph{roof} and \emph{sofa}, as negatives. On the contrary, our XBM offers both semantically \emph{bicycle}-related images and other samples, \eg \emph{wheel} and \emph{clothes}.
These results clearly demonstrate that the proposed XBM can provide \textbf{diverse}, related, and even fine-grained samples to construct negative pairs. 

Our results confirm that (1) existing pair-based approaches suffer from the problem of lacking informative negative pairs to learn a discriminative model, and (2) our XBM module can significantly strengthen the hard mining ability of pair-based DML in a very simple yet efficient manner. See more examples in Supplementary Materials.

\section{Experiments and Results}

\subsection{Implementation Details}
We follow the standard settings in \cite{lifted-structured-loss,n-pairs,Opitz2017BIERB,Kim_2018_ECCV_ABE} for fair comparison. Specifically, we adopt GoogleNet \cite{inception} as the default backbone network if not mentioned. 
The weights of the backbone were pre-trained on ILSVRC 2012-CLS dataset \cite{ILSVRC15}. A 512-d fully-connected layer with $l_2$ normalization is added after the global pooling layer. The default embedding dimension is set as 512. For all datasets, the input images are first resized to $256 \times 256$, and then cropped to $224 \times 224$. Random crops and random flips are utilized as data augmentation during training. For testing, we only use the single center crop to compute the embedding for each instance as \cite{lifted-structured-loss}.
In all experiments, we use the Adam optimizer \cite{kingma:adam} with $5e^{-4}$ weight decay and the PK sampler (P categories, K samples/category) to construct mini-batches.

\subsection{Datasets}
Our methods are evaluated on three large-scale datasets for few-shot image retrieval. Recall$@k$ is reported. The training and testing protocol follow the standard setups:

\noindent\textbf{Stanford Online Products (SOP)} \cite{lifted-structured-loss} contains 120,053 online product images in 22,634 categories. There are only 2 to 10 images for each category. Following \cite{lifted-structured-loss}, we use 59,551 images (11,318 classes) for training, and 60,502 images (11,316 classes) for testing. 

\noindent\textbf{In-shop Clothes Retrieval (In-shop)} contains 72,712 clothing images of 7,986 classes. Following \cite{DeepFashion}, we use 3,997 classes with 25,882 images as the training set. The test set is partitioned to a query set with 14,218 images of 3,985 classes, and a gallery set having 3,985 classes with 12,612 images. 

\noindent\textbf{PKU VehicleID (VehicleID)} \cite{liu2016deep} contains 221,736 surveillance images of 26,267 vehicle categories, where 13,134 classes (110,178 images) are used for training. Following the test protocol described in \cite{liu2016deep}, evaluation is conducted on a predefined small, medium and large test sets which contain 800 classes (7,332 images), 1600 classes (12,995 images) and 2400 classes (20,038 images) respectively.

\begin{table*}[t]
\tablestyle{5.8pt}{1.1}
  \vspace{0em}
    \begin{tabular}{l|cccc|cccccc|cc|cc|cc}
    \multirow{2}[2]{*}{} & \multicolumn{4}{c|}{\multirow{2}[2]{*}{SOP}} & \multicolumn{6}{c|}{\multirow{2}[2]{*}{In-shop}} & \multicolumn{6}{c}{VehicleID} \\
          & \multicolumn{4}{c|}{}         & \multicolumn{6}{c|}{}                         & \multicolumn{2}{c|}{Small} & \multicolumn{2}{c|}{Medium} & \multicolumn{2}{c}{Large} \\ 
    Recall$@K$ (\%)    & 1     & 10    & 100   & 1000  & 1     & 10    & 20    & 30    & 40    & 50    & 1     & 5     & 1     & 5     & 1     & 5 \\ \shline
    Contrastive & 64.0 & 81.4 & 92.1 & 97.8 & 77.1 & 93.0 & 95.2 & 96.1 & 96.8 & 97.1 & 79.5 & 91.6 & 76.2 & 89.3 & 70.0 & 86.0 \\
    \bf Contrastive w/ M & \bf 77.8 & \bf 89.8 & \bf 95.4 & \bf 98.5 & \bf 89.1 & \bf 97.3 & \bf 98.1 & \bf 98.4 & \bf 98.7 & \bf 98.8 & \bf 94.1 & \bf 96.2 & \bf 93.1 & \bf 95.5 & \bf 92.5 & \bf 95.5 \\ \hline
    Triplet & 61.6 & 80.2 & 91.6 & 97.7 & 79.8 & 94.8 & 96.5 & 97.4 & 97.8 & 98.2 & 86.9 & 94.8 & 84.8 & 93.4 & 79.7 & 91.4 \\
    \bf Triplet w/ M & \bf 74.2 & \bf 87.4 & \bf 94.2 & \bf 98.0 & \bf 82.9 & \bf 95.7 & \bf 96.9 & \bf 97.4 & \bf 97.8 & \bf 98.0 & \bf 93.3 & \bf 95.8 & \bf 92.0 & \bf 95.0 & \bf 91.3 & \bf 94.8 \\ \hline
    MS    & 69.7 & 84.2 & 93.1 & 97.9 & 85.1 & 96.7 & 97.8 & 98.3 & 98.7 & 98.8 & 91.0 & 96.1 & 89.4 & 94.8 & 86.7 & 93.8 \\
    \bf MS w/ M & \bf 76.2 & \bf 89.3 & \bf 95.4 & \bf 98.6 & \bf 87.1 & \bf 97.1 & \bf 98.0 & \bf 98.4 & \bf 98.7 & \bf 98.9 & \bf 94.1 & \bf 96.7 & \bf 93.0 & \bf 95.8 & \bf 92.1 & \bf 95.6 \\
    \end{tabular}%
    \vspace{.3em}
  
  \caption{Retrieval results of memory augmented (`w/ M') pair-based methods compared with their respective baselines on three datasets.}
  \label{loss-table}%
  \vspace{-1.em}
\end{table*}%

\subsection{Ablation Study}
We provide ablation study on \textbf{SOP} dataset with \textbf{GoogleNet} to verify the effectiveness of our XBM module.

\noindent\textbf{Memory Ratio.}
 The search space of our cross-batch hard mining can be dynamically controlled by memory ratio $R_{\sM}$. We illustrate the impact of memory ratio to XBM augmented contrastive loss on three benchmarks (in Figure~\ref{fig:bs-mem}, right).
 Firstly, our method significantly outperforms the baseline (with $R_{\sM}=0$), \emph{with over 20\% improvements} on all three datasets using various configurations of $R_{\sM}$. Secondly, our method with mini-batch of 16 can achieve better performance than the non-memory counterpart using 256 mini-batch, \eg with an improvement of 71.7\%$\rightarrow$78.2\% on recall@1, while saving GPU memory considerably. 
 
More importantly, our XBM can boost the contrastive loss largely with small $R_{\sM}$ (\eg on In-shop, \textbf{52.0\%$\rightarrow$ 79.4\% on recall@1 with $R_{\sM}=0.01$}) and its performance is going to be \textbf{saturated} when the memory expands to a moderate size.
 It makes sense, since the memory with a small $R_{\sM}$ (\eg 1\%) already contains thousands of embeddings to generate sufficient valid negative instances on large-scale datasets, especially fine-grained ones, such as In-shop or VehicleID. Therefore, our memory scheme can have consistent and stable performance improvements with a wide range of memory ratios. \\
 
\begin{figure}[t]
\vspace{-0.5em}
\centering
\includegraphics[width=0.47\textwidth]{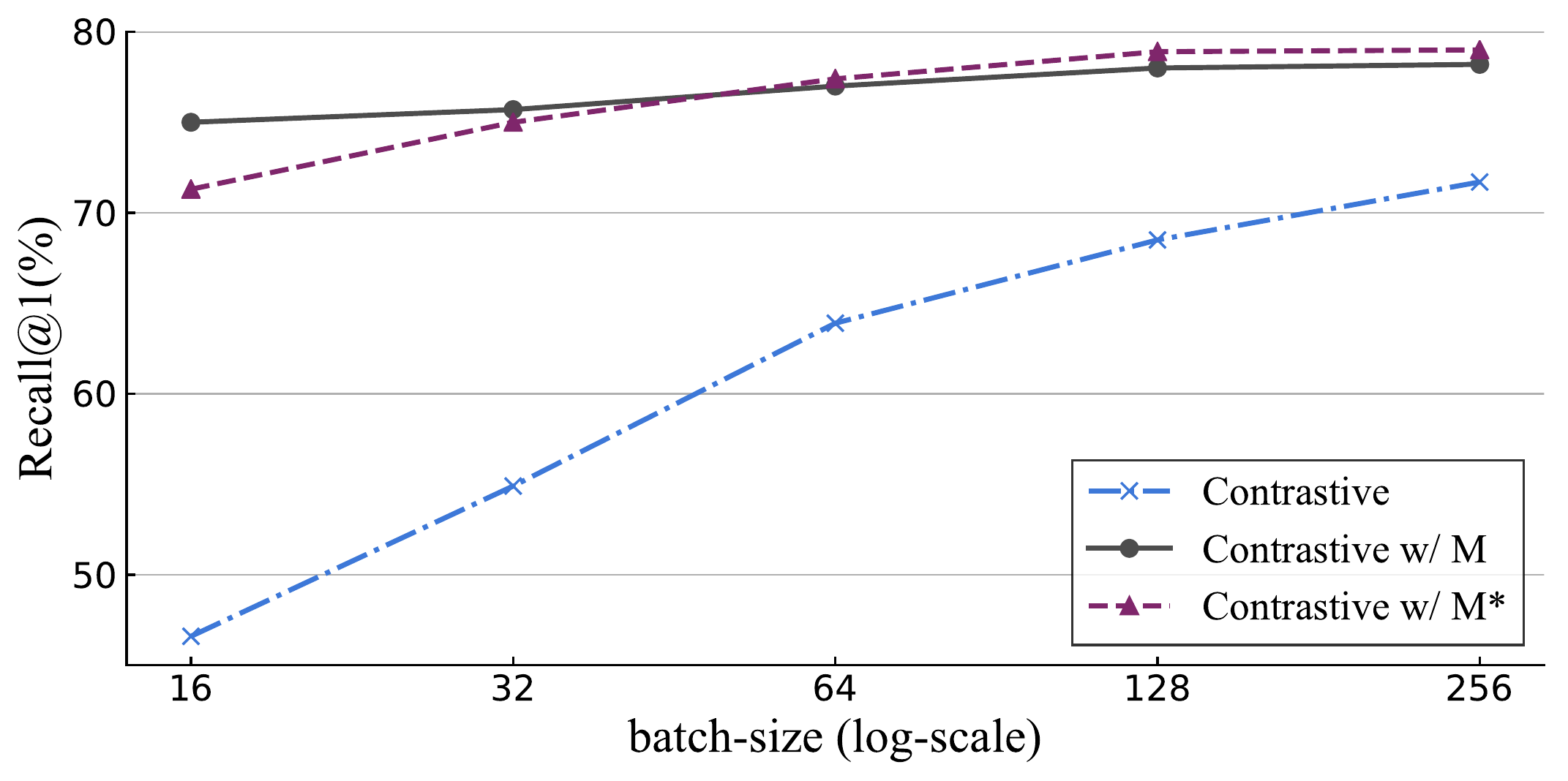}
\caption{\small Performance of contrastive loss by training with different mini-batch sizes. Unlike conventional pair-based methods, XBM augmented contrastive loss is equally effective under \textbf{random shuffle} mini-batch sampler (denoted with superscript *).}
\label{fig:memory_batchsize}
\vspace{-1.5em}
\end{figure}

\noindent\textbf{Mini-batch Size.}
Mini-batch size is critical to the performance of many pair-based approaches (Figure~\ref{fig:bs-mem}, left). We further investigate its impact to our memory augmented pair-based methods (shown in Figure~\ref{fig:memory_batchsize}). Our method has a 3.2\% performance gain by increasing its mini-batch size from 16 to 256, while the original contrastive method has a significantly larger improvement of 25.1\%. Obviously, with the proposed memory module, the impact of mini-batch size is reduced significantly. This indicates that the effect of mini-batch size can be strongly compensated by our memory module, which provides a more principled solution to address the hard mining problem in DML. \\

\noindent\textbf{With General Pair-based DML.}
Our memory module can be directly applied to the GPW framework. We evaluate it with contrastive loss, triplet loss and MS loss.  As shown in Table~\ref{loss-table}, our memory module can improve the original DML approaches significantly and consistently on all benchmarks. Specifically, the memory module remarkably boosts the performance of contrastive loss by \textbf{64.0\%$\rightarrow$77.8\%} and MS loss by 69.7\%$\rightarrow$76.2\%. Furthermore, with sophisticated sampling and weighting approach, MS loss has 16.7\% recall@1 performance improvement over contrastive loss on VehicleID \emph{Large} test set. Such a large gap can be simply filled by our memory module, with \textbf{a further 5.8\% improvement}. MS loss has a smaller improvement because it weights extremely hard negatives heavily which might be outliers, while such a harmful influence is weakened by the equally weighting scheme of contrastive loss. For a detailed analysis see Supplementary Materials (SM).

The results suggest that (1) both straightforward (\eg contrastive loss) and carefully designed weighting scheme (\eg MS loss) can be improved largely by our memory module, and (2) \emph{with our memory module, a simple pair-weighting method (\eg contrastive loss) can easily outperform the most sophisticated, state-of-the-art methods such as MS loss \cite{wang2019multi} by a large margin.} \\

\begin{table}[t]
\vspace{-.5em}
  \tablestyle{7pt}{1.1}
    \begin{tabular}{l|c|c|c|c}
    Method & Time & GPU Mem. & R@1 & Gain \\  \shline
    \textcolor{gray}{Cont. bs. 64} & \textcolor{gray}{2.10 h.} & \textcolor{gray}{5.12 GB} & \textcolor{gray}{63.9} & \textcolor{gray}{-} \\ \hline
    Cont. bs. 256 & 4.32 h. & +15.7 GB & 71.7 & +7.8 \\
    Cont. w/ 1\% $R_M$  & 2.48 h. & +0.01 GB & 69.8 & +5.9\\
    \bf{Cont. w/ 100\% $R_M$} & \bf 3.19 h. & \bf +0.20 GB & \bf 77.4 & \bf +13.5 \\
    \end{tabular}%
    \vspace{.3em}
    
  \caption{Training time and GPU memory cost on 64, 256 mini-batch size and 1\%, 100\% memory ratio with 64 mini-batch size.}
  \label{cost-table}
  \vspace{-1.5em}
\end{table}%

\noindent\textbf{Memory and Computational Cost.} 
We analyze the complexity of our XBM module on memory and computational cost.
On memory cost, The XBM module $\sM$ ($\mathcal{O}(DM)$) and affinity matrix $\widetilde{\mS}$ ($\mathcal{O}(mM)$) requires a negligible 0.2 GB GPU memory for caching the whole training set (Table~\ref{cost-table}).
On computational complexity, the cost of $\widetilde{\mS}$ ($\mathcal{O}(mDM)$) increases linearly with memory size $M$. With a GPU implementation, it takes a reasonable 34\% amount of extra training time \wrt the forward and backward procedure.

It is also worth noting that XBM does not act in the inference phase. It only requires 1 hour extra training time and 0.2GB memory, to achieve a surprising \textbf{13.5\% performance gain} by using a single GPU. 
Moreover, our method can be scalable to an extremely large-scale dataset, \eg with \emph{1 billion} samples, since our XBM module can generate a rich set of valid negatives with a small-memory-ratio XBM, which requires acceptable cost.

\subsection{Quantitative and Qualitative Results}
In this section, we compare our XBM augmented contrastive loss with the state-of-the-art DML methods on three benchmarks on image retrieval. Even though our method can reach better performance with a larger mini-batch size (Figure~\ref{fig:memory_batchsize}), we only use 64 mini-batch which can be implemented on a single GPU with ResNet50 \cite{resnet}. 
Since the backbone architecture and embedding dimension can effect the recall metric, we list the results of our method with various configurations for fair comparison in Table~\ref{sop-table},~\ref{Shop-table} and~\ref{pkuvid-table}. See results on more datasets in SM.

The results demonstrate that our XBM module, with a contrastive loss, can surpass the state-of-the-art methods on all datasets by a large margin. On SOP, our method with R$^\text{128}$ outperforms the current state-of-the-art method: MIC \cite{roth2019mic} by 77.2\% $\rightarrow$ 80.6\%. On In-shop, our method with R$^\text{128}$ achieves even higher performance than FastAP \cite{cakir2019deep} with R$^\text{512}$, and improves by 88.2\%$\rightarrow$91.3\% compared to MIC. On VehicleID, our method outperforms existing approaches considerably. For example, on the \emph{large} test dataset, by using a same G$^\text{512}$, it improves the R@1 of recent A-BIER \cite{opitz2018deep} largely  by 81.9\%$\rightarrow$92.5\%. With R$^\text{128}$, our method surpasses the best results by  87\%$\rightarrow$93\%, which is obtained by FastAP \cite{cakir2019deep} using R$^\text{512}$.

\begin{figure}[t]
\vspace{-1.5em}
\centering
\includegraphics[width=0.47\textwidth, trim=90 60 90 30, clip]{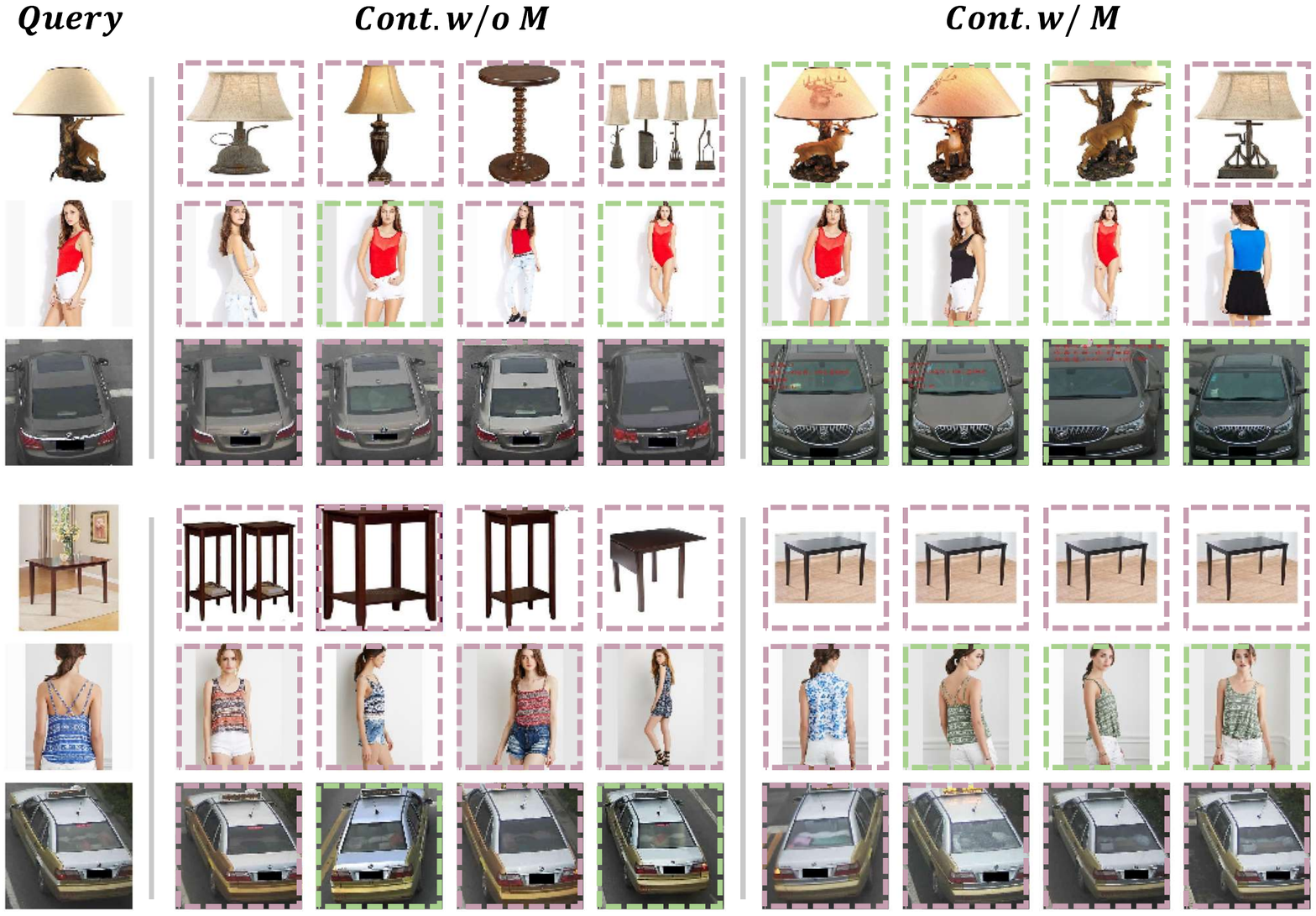}
\caption{\small Top 4 retrieved images w/o and w/ memory module. Correct results are highlighted with {\color[rgb]{0.663, 0.819, 0.556} green}, while incorrect {\color[rgb]{0.729, 0.533, 0.623} purple}.}
\label{fig:ret_imgs}
\vspace{-1.5em}
\end{figure}

Figure~\ref{fig:ret_imgs} shows that our memory module promotes the learning of a more discriminative encoder.
For example, at the first row, our model is aware of \emph{the deer under the lamp} which is a specific character of the query product, and retrieves the correct images. In addition, we also present some bad cases in the bottom rows, where our retrieved results are visually closer to the query than that of baseline model. See more visualizations in SM.
\begin{table}[t]
	\vspace{0em}
	\tablestyle{8pt}{1.1}
		\begin{tabular}{ll|cccc}
			Recall$@K$ (\%)&& 1 & 10 & 100 & 1000\\ \shline
			HDC \cite{hdc} &G$^{\text{384}}$& 69.5 & 84.4 & 92.8 & 97.7 \\
			A-BIER \cite{opitz2018deep} &G$^{\text{512}}$  & 74.2 & 86.9 & 94.0 & 97.8 \\
			ABE \cite{Kim_2018_ECCV_ABE}&G$^{\text{512}}$ &76.3 & 88.4 & 94.8 & 98.2\\
			SM \cite{suh2019stochastic} &G$^{\text{512}}$&75.2 &87.5 &93.7 &97.4\\\hline
			Clustering \cite{struct-clustering} &B$^{\text{64}}$ & 67.0 & 83.7 & 93.2 & -\\
			ProxyNCA \cite{proxyloss} &B$^{\text{64}}$& 73.7 & -& -& -\\
			HTL \cite{HTL} &B$^{\text{512}}$& 74.8& 88.3& 94.8& 98.4\\
			MS \cite{wang2019multi} &B$^{\text{512}}$& 78.2 &90.5 &96.0 &98.7\\
			SoftTriple \cite{qian2019softtriple} &B$^{\text{512}}$& 78.6 & 86.6 & 91.8 & 95.4 \\ \hline
			Margin \cite{sampling} &R$^{\text{128}}$& 72.7 & 86.2 & 93.8 & 98.0\\
			Divide \cite{sanakoyeu2019divide}&R$^{\text{128}}$ & 75.9 & 88.4 & 94.9 & 98.1\\
			FastAP \cite{cakir2019deep} &R$^{\text{128}}$ &73.8 &88.0 &94.9 &98.3\\
			MIC \cite{roth2019mic} &R$^{\text{128}}$ & 77.2 &89.4 &95.6& -\\ \hline\hline
			Cont. w/ M &G$^{\text{512}}$ & 77.4 & 89.6 & 95.4 & 98.4 \\
			Cont. w/ M &B$^{\text{512}}$   & 79.5 & 90.8 & 96.1 & \bf 98.7 \\
			\bf Cont. w/ M &\bf R$^{\text{128}}$   &\bf  80.6 &\bf  91.6 &\bf  96.2 &\bf  98.7 \\
		\end{tabular}
		\vspace{.3em}
		\caption{Recall@$K(\%)$ performance on SOP. `G', `B' and `R' denotes applying GoogleNet, 
		InceptionBN and ResNet50 as backbone respectively, and the superscript is embedding size.}
		\label{sop-table}
	\vspace{-1.5em}
\end{table}

\begin{table}[t]
\small
	\vspace{0em}
		\tablestyle{5pt}{1.1}
		\begin{tabular}{ll|cccccc}
			Recall$@K$ (\%)&& 1 & 10 & 20 & 30 & 40 & 50\\ \shline
			HDC \cite{hdc} & G$^{\text{384}}$& 62.1 & 84.9 & 89.0 & 91.2 & 92.3 & 93.1\\	 
			A-BIER  \cite{opitz2018deep} &G$^{\text{512}}$ & 83.1 & 95.1 & 96.9 & 97.5 & 97.8 & 98.0\\	
			ABE \cite{Kim_2018_ECCV_ABE}  &G$^{\text{512}}$& 87.3 & 96.7 & 97.9 & 98.2 & 98.5 & 98.7\\\hline 
			HTL \cite{HTL} &B$^{\text{512}}$& 80.9& 94.3& 95.8& 97.2& 97.4& 97.8\\
			MS \cite{wang2019multi} &B$^{\text{512}}$& 89.7 &97.9 &98.5 &98.8 &99.1 &99.2\\ \hline
			Divide \cite{sanakoyeu2019divide}  &R$^{\text{128}}$&85.7 &95.5 &96.9 &97.5 &-&98.0 \\
			MIC \cite{roth2019mic} &R$^{\text{128}}$ &88.2 &97.0 &- &98.0 &-&98.8 \\
			FastAP \cite{cakir2019deep} &$R^{\text{512}}$& 90.9 &97.7 &98.5 &98.8 &98.9 &99.1 \\\hline\hline
			Cont. w/ M &G$^{\text{512}}$ & 89.4 & 97.5 & 98.3 & 98.6 & 98.7 & 98.9 \\
			Cont. w/ M &B$^{\text{512}}$ & 89.9 & 97.6 &\bf  98.4 & 98.6 & 98.8 & 98.9 \\
			\bf Cont. w/ M &\bf R$^{\text{128}}$ &\bf  91.3 &\bf  97.8 &\bf  98.4 &\bf  98.7 &\bf  99.0 &\bf  99.1 \\

		\end{tabular}
		\vspace{.3em}
		\caption{Recall@$K(\%)$ performance on In-Shop.}
		\label{Shop-table}
		\vspace{-1em}

\end{table}

\begin{table}[htbp]
\vspace{0em}
  \tablestyle{5pt}{1.1}
    \begin{tabular}{ll|cc|cc|cc} 
    \multicolumn{1}{l}{\multirow{2}[2]{*}{Method}} 
          && \multicolumn{2}{c|}{Small} & \multicolumn{2}{c|}{Medium} & \multicolumn{2}{c}{Large} \\
          && 1   & 5   & 1   & 5   & 1   & 5 \\ \shline
    GS-TRS \cite{em2017incorporating}&  & 75.0  & 83.0  & 74.1  & 82.6  & 73.2  & 81.9  \\
    BIER \cite{Opitz2017BIERB} &G$^{\text{512}}$& 82.6  & 90.6  & 79.3  & 88.3  & 76.0  & 86.4  \\
    A-BIER \cite{opitz2018deep}&G$^{\text{512}}$ & 86.3  & 92.7  & 83.3  & 88.7  & 81.9  & 88.7  \\
    VANet \cite{Chu_2019_ICCV} &G$^{\text{2048}}$ &83.3 &95.9 &81.1 &94.7 &77.2 &92.9\\\hline
    MS  \cite{wang2019multi} &B$^{\text{512}}$& 91.0 & 96.1 & 89.4 & 94.8 & 86.7 & 93.8\\\hline
    Divide \cite{sanakoyeu2019divide}& R$^{\text{128}}$ & 87.7 & 92.9 & 85.7 & 90.4 & 82.9 & 90.2 \\
    MIC \cite{roth2019mic} & R$^{\text{128}}$ &86.9 &93.4& -& - &82.0 &91.0\\
    FastAP \cite{cakir2019deep} &$R^{\text{512}}$ & 91.9 &96.8 &90.6 &95.9 &87.5 &95.1 \\\hline\hline
     Cont. w/ M & G$^{\text{512}}$  &94.0 & 96.3 & 93.2 & 95.4 & 92.5 & 95.5  \\ 
    Cont. w/ M & B$^{\text{512}}$ & 94.6 & \bf  96.9 &  93.4 & \bf 96.0 & \bf 93.0 & \bf 96.1  \\
    \bf Cont. w/ M &\bf R$^{\text{128}}$  & \bf 94.7 & 96.8 & \bf 93.7 & 95.8 & \bf 93.0 & 95.8 \\
    \end{tabular}%
    \vspace{.3em}
    \caption{Recall@$K(\%)$ performance on VehicleID.}
    \label{pkuvid-table}
    \vspace{-1.em}
\end{table}%

\section{Conclusions}
 We have presented a conceptually simple, easy to implement, and memory efficient cross-batch mining mechanism for pair-based DML. In this work, we identify the {``slow drift"} phenomena that the embeddings drift exceptionally slow during the training process. Then we propose a cross-batch memory (XBM) module to dynamically update the embeddings of instances of recent mini-batches, which allows us to collect sufficient hard negative pairs across multiple mini-batches, or even from the whole dataset. Without bells and whistles, the proposed XBM can be directly integrated into a general pair-based DML framework, and improve the performance of existing pair-based methods significantly on image retrieval. In particular, with our XBM, a contrastive loss can easily surpass state-of-the-art methods \cite{wang2019multi, roth2019mic, cakir2019deep} by a large margin on three large-scale datasets. 
 
 This paves a new path in solving for hard negative mining which is a fundamental problem for various computer vision tasks. Furthermore, we hope that the dynamic memory mechanism can be extended to improve a wide variety of machine learning tasks because {\it "slow drift"} is a general phenomenon not only occurring in DML.
\newpage
{\small
\bibliographystyle{ieee_fullname}
\bibliography{egbib}
}
\clearpage
\clearpage


\section*{Supplementary material}

# Cross-Batch Memory for Embedding Learning
# Supplementary Materials

## 1. Results on More Datasets

We further verify the effectiveness of our Cross-Batch Memory (XBM) on three more datasets. CUB-200-2011 (CUB) [11] and Cars-196 (Car) [5] are two widely used fine-grained datasets, which are relatively small. DeepFashion2 [2] is a large-scale dataset just released recently. The details and evaluation protocols of the three datasets are described as below.

**CUB-200-2011 (CUB)** contains 11,788 birds images of 200 classes. There are about 60 images/class. Following [11], we use 5,864 images of 100 classes for training and the remaining 5,924 images for testing.

**Cars-196 (Cars)** contains 16,185 images in 196 model categories, Following [5], we use the first 98 models for training with the rest for testing.

**DeepFashion2** contains 216K clothes images with over 686K commercial-consumer pairs in the training set, whose size is **nearly 7 times** of In-shop. We use ground truth boxes in training and testing. We follow evaluation protocol described in [2], using 10,990 commercial images with 32,550 items as a query set, and 21,438 commercial images with 37,183 items as a gallery set.

**XBM meets Pair-based DML.** We show that XBM consistently improves the performance of various pair-based methods on the three datasets. For instance, by applying our XBM to the conventional contrastive loss, we achieve significant Recall@1 improvements on CUB with +4.4% and Cars with +7.8%, as shown in Table 3. On the large-scale DeepFashion2, XBM has a large improvement of +11.2% as shown in Table 5.

**Comparison with the State-of-the-art.** We compare our method with existing methods in Table 4. On CUB, our XBM with a contractive loss achieves the best recall@1 of 65.8% without any tricks. On Cars, except ABE [4] and A-Bier [8], our XBM augmented contrastive loss reaches the best performance using GoogleNet.

In fact, we observed that several tricks can improve the performance significantly on small-scale datasets. For example, freezing BN layer [12, 9] can increase recall@1 by more than 2% on CUB and Cars, or applying a 10 times smaller learning rate on the pretrained backbones [7]. *However, these tricks show no effect on large-scale datasets* e.g. *SOP, In-shop and VehicleID, which contains sufficient data*

| momentum $m$ | 0 | 0.01 | 0.1 | 0.5 | 0.9 |
|---|---|---|---|---|---|
| SOP | **78.2** | 77.4 | **78.2** | 76.8 | 75.8 |
| In-shop | **89.3** | 88.9 | **89.3** | 87.0 | 83.7 |
| CUB | 60.3 | 60.3 | 60.0 | 60.2 | **61.8** |
| Cars | 78.8 | 79.1 | 79.2 | 78.3 | **80.6** |

Table 1: Recall@1(%) performance of moving average update mechanism with different momentum $m$.

| learning rate $\alpha$ | 0 | 0.01 | 0.1 | 0.5 | 0.9 |
|---|---|---|---|---|---|
| SOP | **78.2** | 77.2 | 77.6 | **78.2** | **78.2** |
| In-shop | **89.3** | 88.7 | 89.0 | 87.8 | 88.4 |
| CUB | 60.3 | 60.2 | **60.7** | 60.0 | 59.4 |
| Cars | 78.8 | 78.8 | **79.3** | 77.9 | 77.8 |

Table 2: Recall@1(%) performance of BP update mechanism with different memory learning rates $\alpha$

*to mitigate overfitting.* Note that to demonstrate the actual effectiveness of our XBM module, the performance of our XBM reported here was trained without such bells and whistles.

## 2. Memory Update

We develop a simple enqueue-dequeue mechanism to update the memory bank of our XBM: enqueue the latest features, and at the same time dequeue the oldest ones. In this experiment, we evaluate two alternative memory update mechanisms: moving average [13, 14] and back-propagation [6], on SOP, In-shop, CUB and Cars datasets with GoogleNet as backbone. Furthermore, we also conduct a faster updated XBM to investigate the effect of feature drift.

**Moving Average Update.** Update a memory embedding $\widetilde{v}_i$ with its current feature $v_i$ as following:

$$\widetilde{v}_i = m\widetilde{v}_i + (1-m)v_i$$
$$\widetilde{v}_i = \widetilde{v}_i/||\widetilde{v}_i||_2,$$

where $m$ is the momentum for the moving average update.

The embeddings in memory is updated slower when the momentum $m$ becomes larger. We study the impact of momentum $m$ to the performance in Table 1. We observed that training with a large momentum can *benefit a small dataset* (*e.g.* CUB or Cars), but may *impair* the performance on a large-scale dataset (*e.g.* SOP or In-shop). It is reasonable

1

| Recall@K(%) | CUB | | | | | | Cars | | | | | |
|---|---|---|---|---|---|---|---|---|---|---|---|---|
| | 1 | 2 | 4 | 8 | 16 | 32 | 1 | 2 | 4 | 8 | 16 | 32 |
| Contrastive | 57.5 | 69.0 | 78.8 | 86.3 | 92.0 | 96.0 | 72.5 | 81.3 | 87.9 | 92.4 | 95.3 | 97.5 |
| **Contrastive w/ M** | **61.9** | **72.9** | **81.2** | **88.6** | **93.5** | **96.5** | **80.3** | **87.1** | **91.9** | **95.1** | **97.3** | **98.2** |
| Triplet | 58.1 | 69.6 | 79.7 | 87.5 | 92.8 | 96.2 | 72.4 | 81.7 | 88.1 | 92.7 | 95.5 | 97.3 |
| **Triplet w/ M** | **60.0** | **71.1** | **80.7** | **88.0** | **93.2** | **96.4** | **78.5** | **86.4** | **91.6** | **94.8** | **96.9** | **98.4** |
| MS | 58.2 | 69.8 | 79.9 | 87.3 | 92.8 | 96.0 | 75.7 | 84.6 | 90.1 | 94.4 | 96.9 | 98.4 |
| **MS w/ M** | **61.8** | **72.3** | **81.5** | **88.5** | **93.0** | **96.1** | **76.5** | **84.1** | **90.0** | **93.8** | **96.3** | **98.0** |

Table 3: Retrieval results of XBM augmented ('w/ M') pair-based DML methods and baseline methods with GoogleNet on CUB and Cars.

| Recall@$K$ (%) | | CUB | | | | | | Cars | | | | | |
|---|---|---|---|---|---|---|---|---|---|---|---|---|---|
| | | 1 | 2 | 4 | 8 | 16 | 32 | 1 | 2 | 4 | 8 | 16 | 32 |
| Smart Mining [3] | G$^{64}$ | 49.8 | 62.3 | 74.1 | 83.3 | - | - | 64.7 | 76.2 | 84.2 | 90.2 | - | - |
| HDC [10] | G$^{384}$ | 53.6 | 65.7 | 77.0 | 85.6 | 91.5 | 95.5 | 73.7 | 83.2 | 89.5 | 93.8 | 96.7 | 98.4 |
| A-BIER [8] | G$^{512}$ | 57.5 | 68.7 | 78.3 | 86.2 | 91.9 | 95.5 | 82.0 | 89.0 | 93.2 | 96.1 | 97.8 | 98.7 |
| ABE [4] | G$^{512}$ | 60.6 | 71.5 | 79.8 | 87.4 | - | - | 85.2 | 90.5 | 94.0 | 96.1 | - | - |
| Clustering [10] | B$^{64}$ | 48.2 | 61.4 | 71.8 | 81.9 | - | - | 58.1 | 70.6 | 80.3 | 87.8 | - | - |
| ProxyNCA [6] | B$^{64}$ | 49.2 | 61.9 | 67.9 | 72.4 | - | - | 73.2 | 82.4 | 86.4 | 87.8 | - | - |
| HTL [1] | B$^{512}$ | 57.1 | 68.8 | 78.7 | 86.5 | 92.5 | 95.5 | 81.4 | 88.0 | 92.7 | 95.7 | 97.4 | 99.0 |
| MS [12] | B$^{512}$ | 65.7 | **77.0** | **86.3** | **91.2** | **95.0** | **97.3** | 84.1 | 90.4 | 94.0 | 96.5 | **98.0** | **98.9** |
| SoftTriple [9] | B$^{512}$ | 65.4 | 76.4 | 84.5 | 90.4 | - | - | **84.5** | **90.7** | **94.5** | **96.9** | - | - |
| Contrative w/ M | G$^{512}$ | 61.9 | 72.9 | 81.2 | 88.6 | 93.5 | 96.5 | 80.3 | 87.1 | 91.9 | 95.1 | 97.3 | 98.2 |
| Contrative w/ M | B$^{512}$ | **65.8** | 75.9 | 84.0 | 89.9 | 94.3 | 97.0 | 82.0 | 88.7 | 93.1 | 96.1 | 97.6 | 98.6 |

Table 4: Recall@$K$(%) performance on CUB and Cars.

| Recall@K(%) | | 1 | 10 | 20 |
|---|---|---|---|---|
| Match RCNN [2] | | 26.8 | 57.4 | 66.5 |
| Contrative | G$^{512}$ | 29.3 | 51.9 | 60.3 |
| Contrative w/ M | G$^{512}$ | 40.5 | 63.2 | 69.4 |
| Contrative w/ M | B$^{512}$ | 40.9 | 63.3 | 69.6 |
| **Contrative w/ M** | **R$^{128}$** | **41.9** | **64.6** | **70.7** |

Table 5: Recall@$K$(%) performance on DeepFashion2.

| Recall@$k$(%) | 1 | 10 | 100 | 1000 |
|---|---|---|---|---|
| update $\times 1$ | 77.4 | 89.6 | 95.4 | 98.4 |
| update $\times 10$ | 77.4 | 89.8 | 95.5 | 98.5 |

Table 6: Recall@$k$(%) performance on SOP with update $\times 1$ and $\times 10$ configurations.

because on a small-scale dataset, the training epoch is short (*e.g.* 100 iters.), which enables a small feature drift between adjacent epochs, while a large-scale dataset has a longer epoch, and the features computed at the past epochs are highly possible to be out-of-date, and thus a large momentum may hinder the training process. Moreover, the moving average update may benefit the training by enhancing the embeddings in the memory by *aggregating its embeddings of an instance with different augmentations* when the feature drift is small.

**Back-Propagation (BP) Update.** Besides substituting the memory features of instances sampled into current mini-batch, BP method updates the memory features of each instance based on its gradients computed at back-propagation (BP):

$$\widetilde{v}_i = \widetilde{v}_i + \alpha \frac{\partial \widetilde{\mathcal{L}}}{\partial \widetilde{v}_i}$$
$$\widetilde{v}_i = \widetilde{v}_i / ||\widetilde{v}_i||_2,$$

where $\alpha$ is the learning rate of memory features.

In BP update, the memory embeddings are optimized along with the model, and serve as proxies in proxy-based DML methods [6, 9]. Obviously, BP requires much more memory and computational cost to compute and save the gradients compared to our XBM. However, it cannot obtain clear performance improvements in all datasets as shown in Table 2. This suggests that the embeddings drift slowly, and the past mini-batches can largely represent the distribution of current embedding space.

**Faster Update.** Generally, we update the XBM with **one mini-batch** at each iteration ($\times 1$). To further evaluate the side effect of feature drift in our XBM, at each iteration, ten mini-batches ($\times 10$) are enqueued into the XBM memory queue. This accelerates XBM update 10 times faster, which makes the feature drift smaller. As shown in Table 6, the $\times 10$ update cannot bring clear performance gain, which suggests that the natural feature drift is slow enough and cannot hinder the performance.

## 3. Feature Drift on General Tasks

*"Slow drift"* phenomena not only exist in pair-based DML, but also happen in other machine learning tasks, *e.g.* image recognition. Figure 1 illustrates that the normalized

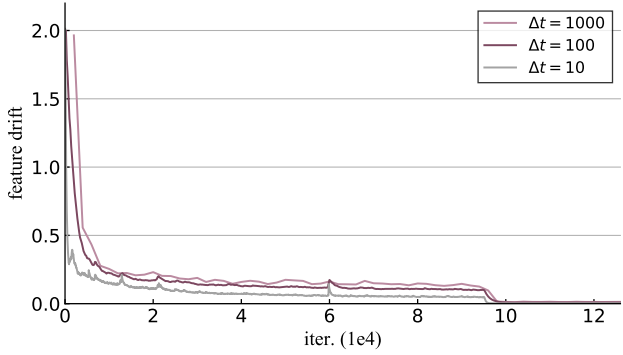

Figure 1: **Feature drift** with different steps of ResNet50 on ImageNet trained *from scratch*. The embeddings drift within a relatively small distance even under a large interval, *e.g.* $\Delta t = 1000$.

embeddings at global pooling layer of ResNet50 drift at a slow rate when trained with cross entropy loss on ImageNet dataset.

## 4. Hyperparameters

In Table 7, we list all the key hyperparameters applied in our experiments. Our XBM achieves outstanding performance on large-scale datasets and comparable results on small-scale datasets without any training trick or large training iterations.

|  | init. lr. | lr.$\times 0.1$ | total iter. | $R_\mathbb{M}(\%)$ | $\alpha$ | $m$ |
|---|---|---|---|---|---|---|
| SOP | 3e-4 | 24k | 34k | 1 | 0 | 0 |
| In-shop | 3e-4 | 24k | 34k | 0.2 | 0 | 0 |
| VehicleID | 1e-4 | 30k | 50k | 0.5 | 0 | 0 |
| DeepFashion2 | 3e-4 | 20k | 36k | 1 | 0 | 0 |
| CUB | 3e-5 | - | 1.4k | - | 0.2 | 0.9 |
| Cars | 1e-4 | 1.4k | 2k | - | 0.1 | 0.9 |

Table 7: **Hyperparameters** used in training memory augmented models to compare with state-of-the-art.

## 5. Proof of Lemma 1

*Proof.* The gradient of the accurate loss and the approximated loss can be computed as below:

$$\frac{\partial \mathcal{L}}{\partial \boldsymbol{\theta}} = \frac{\partial \mathcal{L}}{\partial \boldsymbol{v}_i} \frac{\partial \boldsymbol{v}_i}{\partial \boldsymbol{\theta}} = \boldsymbol{v}_j \frac{\partial \boldsymbol{v}_i}{\partial \boldsymbol{\theta}}$$
$$\frac{\partial \widetilde{\mathcal{L}}}{\partial \boldsymbol{\theta}} = \frac{\partial \widetilde{\mathcal{L}}}{\partial \boldsymbol{v}_i} \frac{\partial \boldsymbol{v}_i}{\partial \boldsymbol{\theta}} = \widetilde{\boldsymbol{v}}_j \frac{\partial \boldsymbol{v}_i}{\partial \boldsymbol{\theta}}$$

Then, the gradient error is:

$$\begin{aligned}\left\|\frac{\partial \mathcal{L}}{\partial \boldsymbol{\theta}} - \frac{\partial \widetilde{\mathcal{L}}}{\partial \boldsymbol{\theta}}\right\|_2^2 &= \left\|(\boldsymbol{v}_j - \widetilde{\boldsymbol{v}}_j)\frac{\partial \boldsymbol{v}_i}{\partial \boldsymbol{\theta}}\right\|_2^2 \\ &\leq \left\|\boldsymbol{v}_j - \widetilde{\boldsymbol{v}}_j\right\|_2^2 \left\|\frac{\partial \boldsymbol{v}_i}{\partial \boldsymbol{\theta}}\right\|_2^2 \\ &\leq \left\|\frac{\partial f(\boldsymbol{x}_i; \boldsymbol{\theta}^t)}{\partial \boldsymbol{\theta}}\right\|_2^2 \epsilon \\ &\leq C\epsilon.\end{aligned} \quad (1)$$

□

Empirically, we observed that $\left\|\frac{\partial f(\boldsymbol{x}_i; \boldsymbol{\theta}^t)}{\partial \boldsymbol{\theta}}\right\|_2^2$ is usually less than 1 so that the gradient error can be strictly controlled by the small feature drift.

## 6. Visualization.

We visualize a number of samples to investigate the performance of our XBM, including the mined negatives in training and the retrieved examples in testing. All examples are *randomly* selected from SOP, In-shop and VehicleID by following some rules.

Figure 2 demonstrates the hard negatives mined from memory with over 0.5 similarities. The negatives in each row are uniformly sampled from a sequence sorted by similarities. These results clearly demonstrate that the proposed XBM can provide **diverse**, visually related, and even fine-grained samples to construct informative negative pairs.

Furthermore, we select the anchors having a similarity with the hardest samples over 0.8, and show the top 10 negatives in Figure 3. Some of the presented negatives are extremely similar or even exactly the same items with corresponding anchor images.

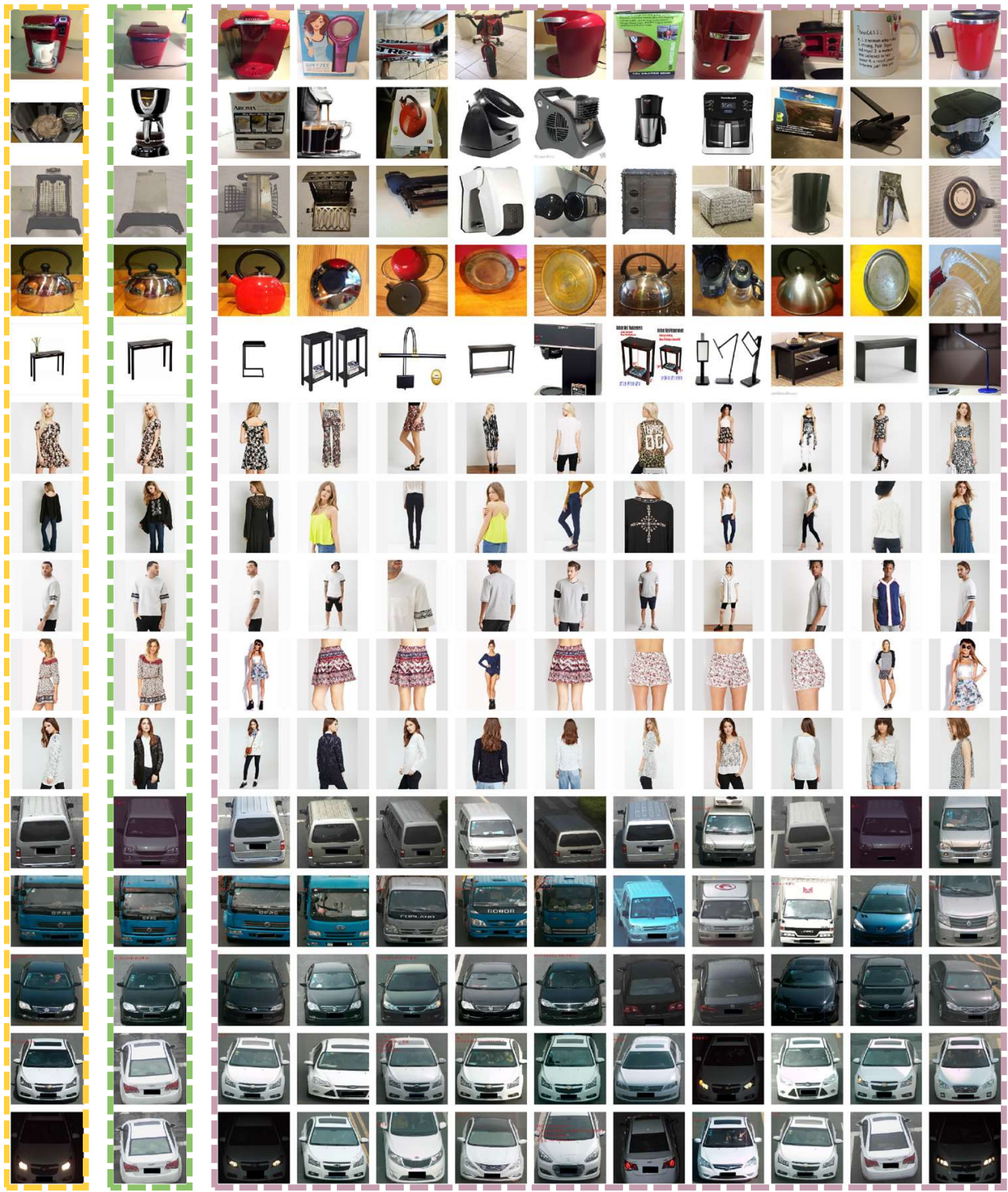

Figure 2: Given an anchor image (yellow), we present examples of a positive (green), and multiple negatives mined from memory (purple) uniformly sorted from hard to simple. The demonstrated anchors are *randomly* chosen from training datasets.

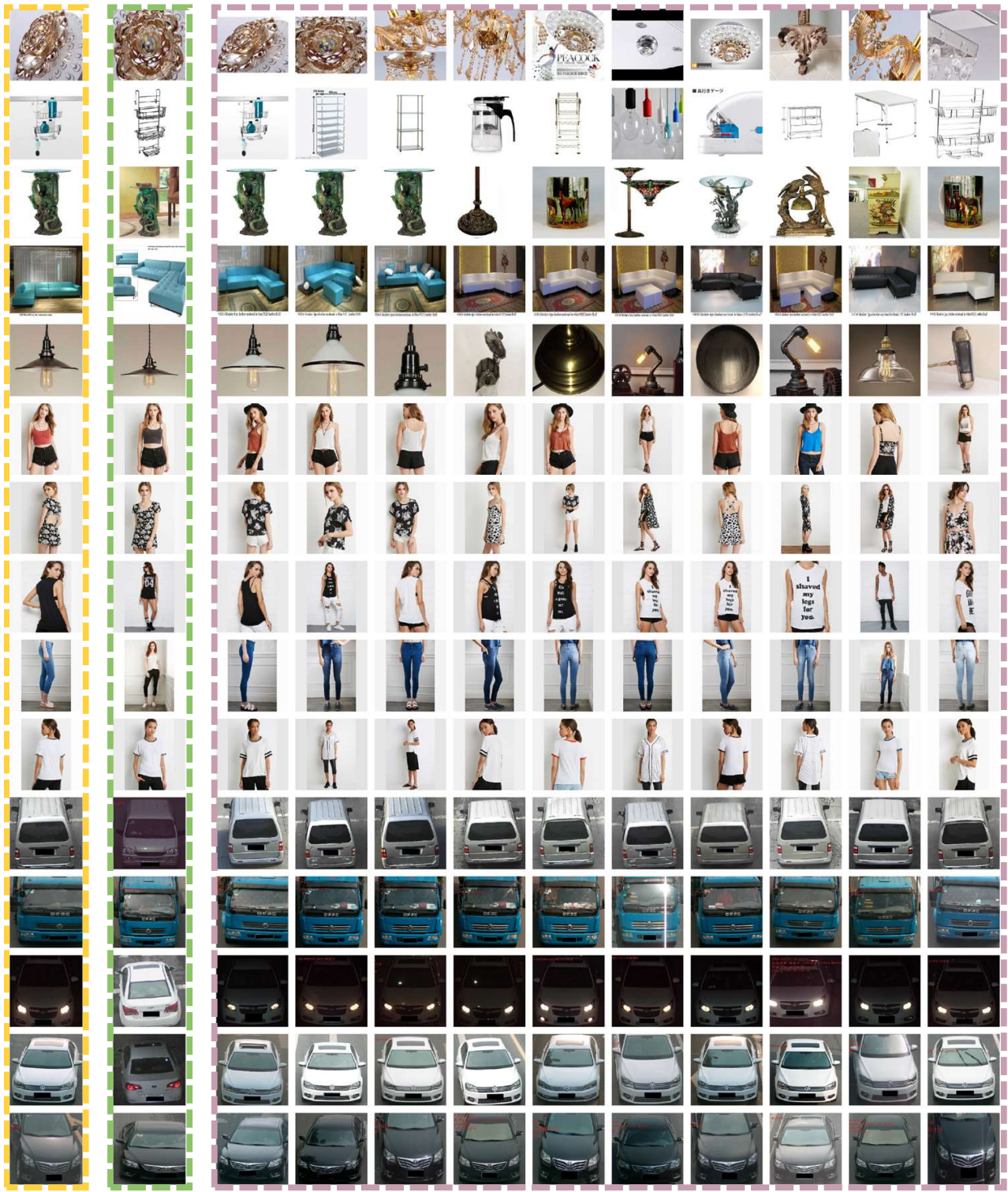

Figure 3: Given an anchor image (yellow), we present examples of a positive (green), and top 10 negatives mined from memory (purple). The examples are *randomly* chosen from anchors whose top 1 negative has over 0.8 similarity.

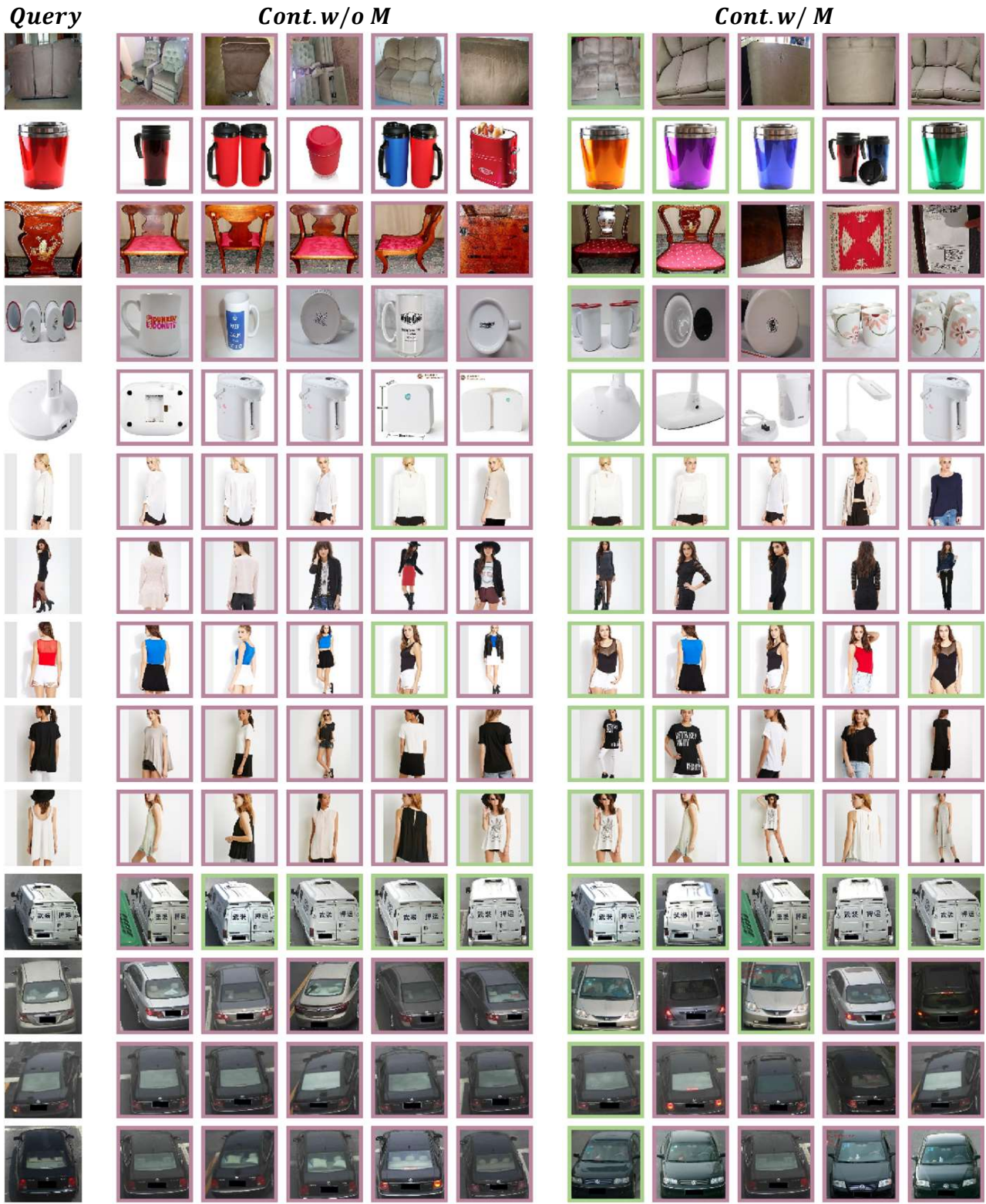

Figure 4: Top 5 retrieved images w/o and w/ memory module. We randomly select the examples with the correct top 1 predictions given by our XBM but incorrect by the baseline model. Correct results are highlighted with green, while incorrect purple.

Figure 5: Top 5 retrieved images w/o and w/ memory module. We randomly select the examples with the wrong top 1 predictions by both the baseline model and our XBM. Correct results are highlighted with green, while incorrect purple.



\end{document}